\documentclass{article} 

\usepackage{iclr2027_conference,times}
\iclrfinalcopy 

\usepackage[T1]{fontenc}
\usepackage{hyperref}
\usepackage{url}
\usepackage{graphicx}
\usepackage{booktabs}
\usepackage{amsmath}
\usepackage{amssymb}
\usepackage{microtype}
\usepackage{xspace}
\usepackage[textsize=footnotesize]{todonotes}
\usepackage{tabularx}
\usepackage{subcaption}
\usepackage{wrapfig}
\usepackage{multirow}
\usepackage{tikz}
\usetikzlibrary{positioning, arrows.meta, fit, backgrounds, calc, shapes.geometric, decorations.pathreplacing}

\newcommand{\modelname}{SceneLM\xspace}
\newcommand{\pipelinename}{OVAL\xspace}
\newcommand{\add}{\texttt{ADD}\xspace}
\newcommand{\edit}{\texttt{EDIT}\xspace}
\newcommand{\remove}{\texttt{REMOVE}\xspace}

\newcommand{\sparse}{Sparse\xspace}
\newcommand{\dense}{Dense\xspace}
\newcommand{\pipelinefullname}{Open-Vocabulary Automatic Labeling\xspace}

\newcommand{\parsection}[1]{\noindent\textbf{#1:}}

\usepackage[capitalize]{cleveref}
\crefname{section}{Sec.}{Secs.}
\Crefname{section}{Section}{Sections}
\Crefname{table}{Table}{Tables}
\crefname{table}{Tab.}{Tabs.}

\usepackage{colortbl} 
\definecolor{CornflowerBlue}{RGB}{100, 149, 237}
\definecolor{YellowGreen}{RGB}{154, 205, 50}
\definecolor{Apricot}{RGB}{251, 206, 177}
\definecolor{Red}{rgb}{1, 0.7, 0.7} 
\definecolor{Orange}{rgb}{1, 0.85, 0.7} 
\definecolor{Yellow}{rgb}{1, 1, 0.7} 
\def\bestrow{\rowcolor{gray!20}}

\title{A Scene Language Model for Open-Vocabulary Scene Mapping}

\author{
\begin{tabular}{c}
Adam Lilja$^{1,2}$ \quad
Fabio H\"ubel$^{3}$ \quad
Siming He$^{4}$ \quad
Junsheng Fu$^{2}$ \quad
Claire Tomlin$^{4}$ \\
Lars Hammarstrand$^{1}$ \quad
Jitendra Malik$^{4}$ \quad
Jonas Frey$^{4}$ \quad
Marco Pavone$^{3,5}$ 
\\
\\[0.2em]
$^{1}$ Chalmers \qquad
$^{2}$ Zenseact \qquad
$^{3}$ Stanford \qquad
$^{4}$ UC Berkeley \qquad
$^{5}$ NVIDIA
\end{tabular}
}

\begin{document}
\maketitle

%
%
\begin{abstract}
\vspace{-16pt}
Open-vocabulary 3D scene mapping aims to build a persistent representation of the objects in an environment.
Existing systems typically rely on engineered mapping pipelines to associate observations, merge information across views, and maintain a consistent scene representation over time.
Many additionally store feature-rich object representations, such as embeddings or image crops, increasing the size and complexity of the persistent memory.
We introduce \modelname, a \textbf{Scene}--\textbf{L}anguage \textbf{M}odel that directly maintains a textual scene map.
The full scene is represented as a structured text list of objects, which serves as the model's only persistent memory.
For each input image, the model reads the current scene state and updates the map by adding, editing, and removing objects.
To learn this behavior, we introduce supervision tasks for iterative scene map maintenance together with an automatic annotation pipeline that generates training data from images without human labels.
We evaluate \modelname on both a language-grounded retrieval benchmark and a localization benchmark.
Across both benchmarks, the model produces a scene map that achieves competitive performance with complete mapping systems built from dedicated perception and geometric modules while producing a scene representation that is $6$--$12\times$ more compact.
We further show that \modelname can be run online on an edge device through experiments on a quadruped.
These results show that a persistent open-vocabulary 3D scene map can be maintained directly by a single vision--language model using only a lightweight text representation.
Training and inference code is available on the
\href{https://goldengait.github.io/scenelm/}{Project page}.
\end{abstract}


\begin{figure}[h!]
   \centering
    \includegraphics[width=0.9\linewidth]{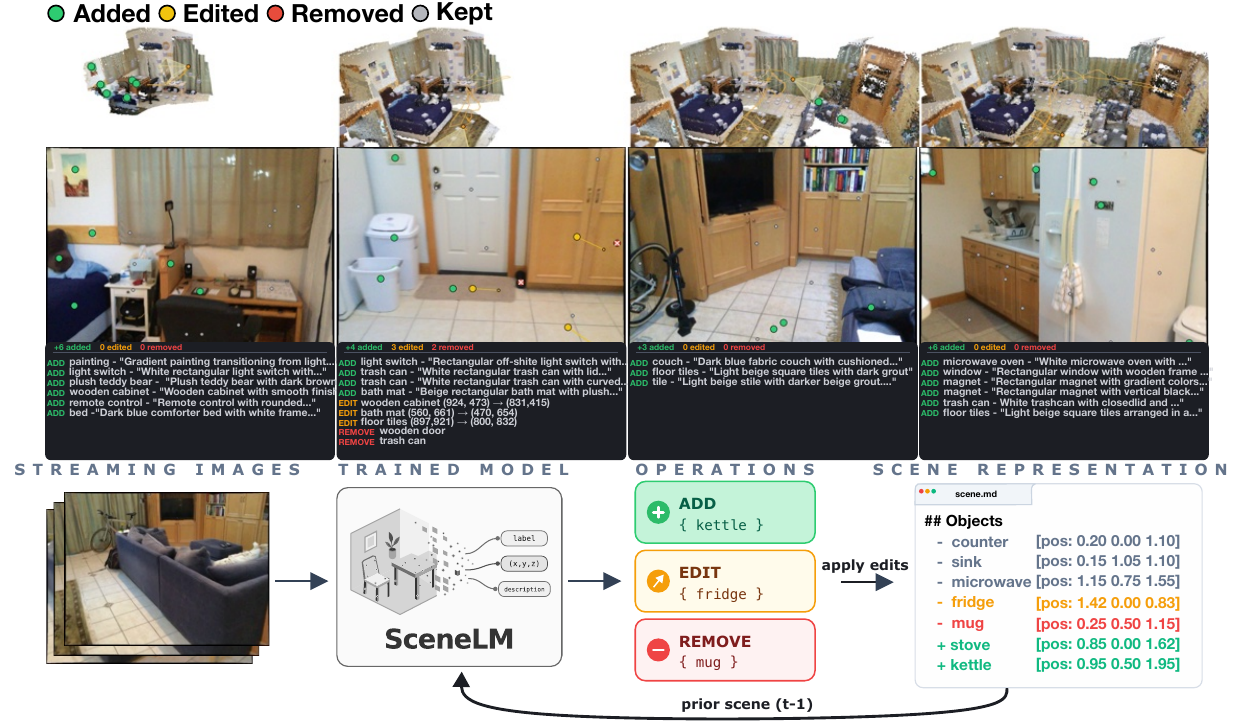}
    \caption{
    Overview of \modelname. 
    Given an RGB observation and the current textual scene map, \modelname predicts \add, \edit, and \remove operations that update the persistent scene map.
    The colored point clouds in the top row are for illustration and are obtained by aggregating the posed RGB-D observations. 
    The four example observations in the middle row illustrate how the textual scene map is incrementally updated as objects are added, edited, and removed. 
    }
    \label{fig:scenelm}
\end{figure}

\section{Introduction}

A robot must maintain a persistent representation of its environment to remember which objects have been observed, where they are located, and how they relate across observations.
In open-vocabulary settings, language provides a natural interface for referring to and querying objects in this memory.
Such persistent scene understanding is important for navigation, manipulation, and long-horizon embodied interaction.
Current open-vocabulary mapping systems typically construct this memory using explicit pipelines that project detections into 3D, associate observations across views, and maintain object tracks or scene graphs~\citep{he2026farm,gorlo2025daaam}.
Many additionally store visual embeddings, image crops, or segmentation masks, resulting in increasingly complex and memory-intensive scene representations.
Rather than extending these pipelines with further specialized components, we ask whether scene-map maintenance itself can be learned by a single vision--language model using only a lightweight textual representation as persistent memory.

We investigate this question with \modelname, an autoregressive vision--language model that maintains a structured text list of objects, locations, and short descriptions.
We call this formulation a \emph{Scene Language Model}: a vision--language model that maintains and updates a persistent scene state through language.
Given a posed RGB-D observation and the current scene map, \modelname predicts \add, \edit, and \remove operations that update this state (\cref{fig:scenelm}).
The model therefore learns operations normally handled by an explicit mapping backend, including discovering objects, associating observations across views, correcting existing entries, and removing false or duplicate objects.
No visual embeddings, image crops, or feature banks are retained between observations.

Learning these behaviors requires supervision for how a scene map changes as new observations arrive.
We introduce training tasks for object insertion, association, correction, and removal, together with \pipelinename (\pipelinefullname), an automatic annotation pipeline that generates object locations, noun phrases, and short descriptions from images and filters low-quality annotations.
This allows map-maintenance supervision to be generated from individual images rather than manually annotated 3D scene maps, while the trained model is applied iteratively to streaming observations.

We evaluate \modelname on complementary open-vocabulary mapping benchmarks for language-grounded retrieval~\citep{he2026farm} and 3D object localization for navigation~\citep{zhang2024tagmap}.
Across both benchmarks, \modelname matches or exceeds several complete mapping systems built from dedicated perception and geometric modules, while using a $6$--$12\times$ smaller scene representation.

Our contributions are:
\begin{enumerate}
    \item We introduce \modelname, a Scene Language Model maintaining a persistent open-vocabulary scene map through \add, \edit, and \remove operations over a structured textual state.
    \item We introduce map-maintenance supervision tasks and the \pipelinename automatic annotation pipeline, enabling these behaviors to be learned from automatically labeled individual images rather than manually annotated 3D scene maps.
    \item We show that the resulting text-based scene memory achieves competitive retrieval and localization performance while being more compact than feature-rich mapping representations.
\end{enumerate}

\section{Related Work}
\label{sec:related}

\textbf{Open-vocabulary 3D scene mapping}
typically combines learned perception with an explicit geometric backend that maintains a persistent scene representation.
ConceptFusion~\citep{jatavallabhula2023conceptfusion} fuses dense visual features into a point cloud, while ConceptGraphs~\citep{gu2024conceptgraphs} constructs object-centric scene graphs through cross-view geometric association.
Later work adds hierarchical representations~\citep{werby2024hovsg}, task-aware abstractions~\citep{maggio2024clio}, and long-term temporal consistency~\citep{gorlo2025daaam}; TagMap~\citep{zhang2024tagmap}, OpenScene~\citep{peng2023openscene}, 
FARM~\citep{he2026farm}, and BBQ~\citep{linok2025beyond} follow the same general pattern of open-vocabulary perception coupled with an engineered mapping backend.
In contrast, \modelname maintains a structured textual scene state directly with a single vision--language model.
Our aim is to replace specialized association and map-maintenance components with a learned process that jointly handles perception, object association, and scene-state updates.

\textbf{Vision--language models for 3D reasoning}
such as 3D-LLM~\citep{hong20233dllm}, LLaVA-3D~\citep{zhu2024llava3d}, and Video-3D LLM~\citep{zheng2025video3dllm} combine visual and geometric information for tasks including grounding, question answering, and instruction following.
These methods generally assume that a scene representation has already been constructed, whereas \modelname incrementally builds and maintains that representation itself.

\textbf{Automatic supervision for scene understanding}
uses foundation models to generate spatial-language supervision without manual 3D annotation.
SpatialVLM~\citep{chen2024spatialvlm}, 3D-LLM~\citep{hong20233dllm}, and ReMEmbR~\citep{anwar2025remembr} construct such supervision from images, videos, or reconstructed environments.
Our annotation pipeline follows this direction but targets scene-map maintenance rather than question answering or scene description, producing supervision for object insertion, cross-view association, correction, and removal.
\begin{figure}[t]
  \centering
  \includegraphics[width=1.0\linewidth]{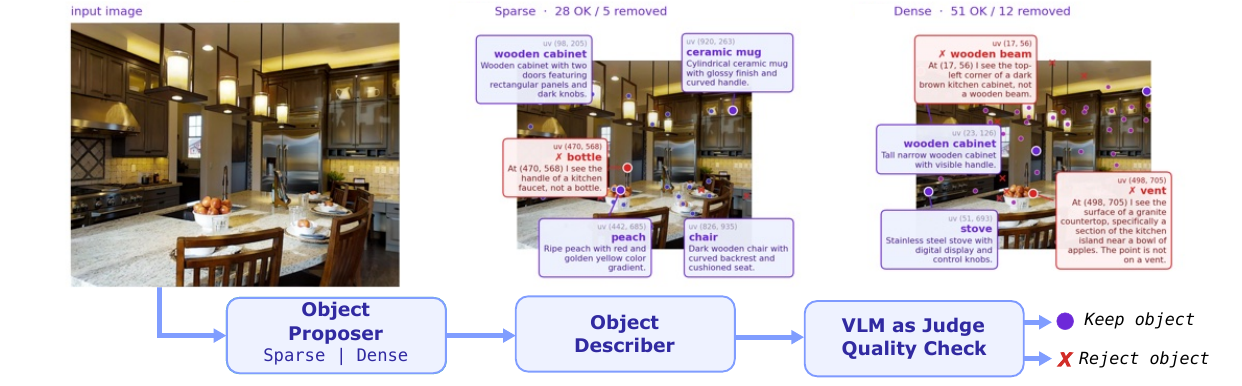}
  \caption{\textbf{The \pipelinename automatic labeling pipeline.} Given an input image, \pipelinename generates object proposals using a sparse open-vocabulary detector or dense promptable segmentor, then assigns each object a 2D anchor point, noun phrase, and short description. Dots indicate object locations and highlighted examples show their annotations. A VLM verifies each annotation for visual and semantic consistency; rejected objects are marked with a red \texttimes ~and accompanied by the rejection reason. Accepted and rejected annotations are retained to construct training supervision.
  }
  \label{fig:pipeline}
\end{figure}

\section{The \pipelinename data pipeline}
\label{sec:method-pipeline}

Manual 3D scene-map annotation is expensive and difficult to scale.
We therefore introduce \pipelinename (\textbf{O}pen-\textbf{V}ocabulary \textbf{A}utomatic \textbf{L}abeling), which uses foundation models to generate object-level pseudo-labels directly from images without human annotation.
For each object, \pipelinename produces a 2D point, noun phrase, and short textual description.
\cref{fig:pipeline} summarizes the pipeline; below we describe its stages and evaluate the resulting label quality.

\subsection{Labeling pipeline}
\label{sec:data-stages}

\textbf{(1) Object proposals:}
We use either a \emph{sparse} open-vocabulary detector or a \emph{dense} promptable segmentor to generate object proposals, denoted \pipelinename-Sparse and \pipelinename-Dense.
The dense variant iteratively samples query points in previously uncovered image regions, generates masks, and repeats until a fixed proposal budget is reached, yielding substantially higher object coverage.
For each proposal, we retain only the most interior mask pixel as a 2D anchor point; masks are used only within the labeling pipeline.

\textbf{(2) Object descriptions:}
A mask-conditioned captioning model generates a long-form description for each proposal, which is condensed into a noun phrase and short description.
Each annotation is therefore represented as a \{2D point, label, short description\} triplet.

\textbf{(3) Quality filtering:}
A VLM evaluates each annotation for relevance and consistency with the visual evidence, removing fine-grained clutter and incorrect labels or descriptions.
Rejected annotations are retained as structured negatives and later used as \remove training candidates.

\textbf{(4) Auxiliary training data:}
To supervise \edit and \remove, we construct controlled corruption pools from the generated annotations.
These include
(i) visually similar hard negatives with different semantics, encouraging fine-grained discrimination;
(ii) synonym-based variants, exposing the model to alternative valid descriptions of the same object; and
(iii) random distractors, creating clearly incorrect object-label pairings.
Together, these corruptions provide structured candidates for learning when an existing map entry should be corrected, preserved, or removed.

\subsection{Implementation details}
In our experiments, \pipelinename-Sparse uses YOLOE~\citep{wang2025yoloe} for object proposals, while \pipelinename-Dense uses SAM~\citep{kirillov2023sam}.
DescribeAnything~\citep{lian2025describeanything} generates mask-conditioned long-form descriptions, which Qwen3~\citep{qwen3vl2025} compresses into a noun phrase and short description.
For quality filtering, Qwen3-VL-30B evaluates each object against the image evidence and produces an accept/reject decision together with a textual justification.
Further implementation details and comparison against human judgments are provided in \cref{app:sec:quality-control}.

\subsection{The dataset}
\label{sec:data-dataset}

\pipelinename is designed to scale across heterogeneous datasets with minimal adaptation.
We apply it to $1.3$M images from indoor navigation and scene-understanding datasets (NaVILA~\citep{cheng2024navila}, HM3D~\citep{ramakrishnan2021hm3d}, ScanNet~\citep{dai2017scannet}, SUN RGB-D~\citep{song2015sunrgbd}, Matterport3D~\citep{chang2017matterport3d}), driving datasets (nuScenes and nuImages~\citep{caesar2020nuscenes}), and web imagery (COCO~\citep{lin2014coco}).
Full dataset details are provided in \cref{app:sec:dataset}.
\pipelinename-\sparse produces $10.7$M annotations, averaging $8.2$ objects per image, of which $81\%$ pass the VLM quality filter.
\pipelinename-\dense produces $41.6$M annotations, averaging $32$ objects per image, with $65\%$ retained.
Each annotation consists of a 2D point, noun phrase, and short description.
Unless stated otherwise, we train \modelname using \pipelinename-\sparse.

We manually assess pseudo-label quality on 100 varied images.
An annotation is marked \emph{good} only if its 2D point, noun phrase, and short description are all correct and visually supported.
Under this criterion, $85\%$ of \sparse and $69\%$ of \dense annotations are judged good.
The VLM quality filter agrees with human judgments on approximately $80\%$ of annotations (Cohen's $\kappa=0.47$, balanced accuracy $0.74$), and $84\%$ of the retained annotations are human-verified as correct.
Further analysis is provided in \cref{app:sec:quality-control}.
The generated annotations form a curated pool rather than direct training targets.
Both accepted and rejected annotations are retained and sampled to construct the different map-maintenance objectives.

\begin{figure*}[t]
    \centering

    \begin{subfigure}[t]{0.32\textwidth}
        \centering
        \includegraphics[width=\linewidth]{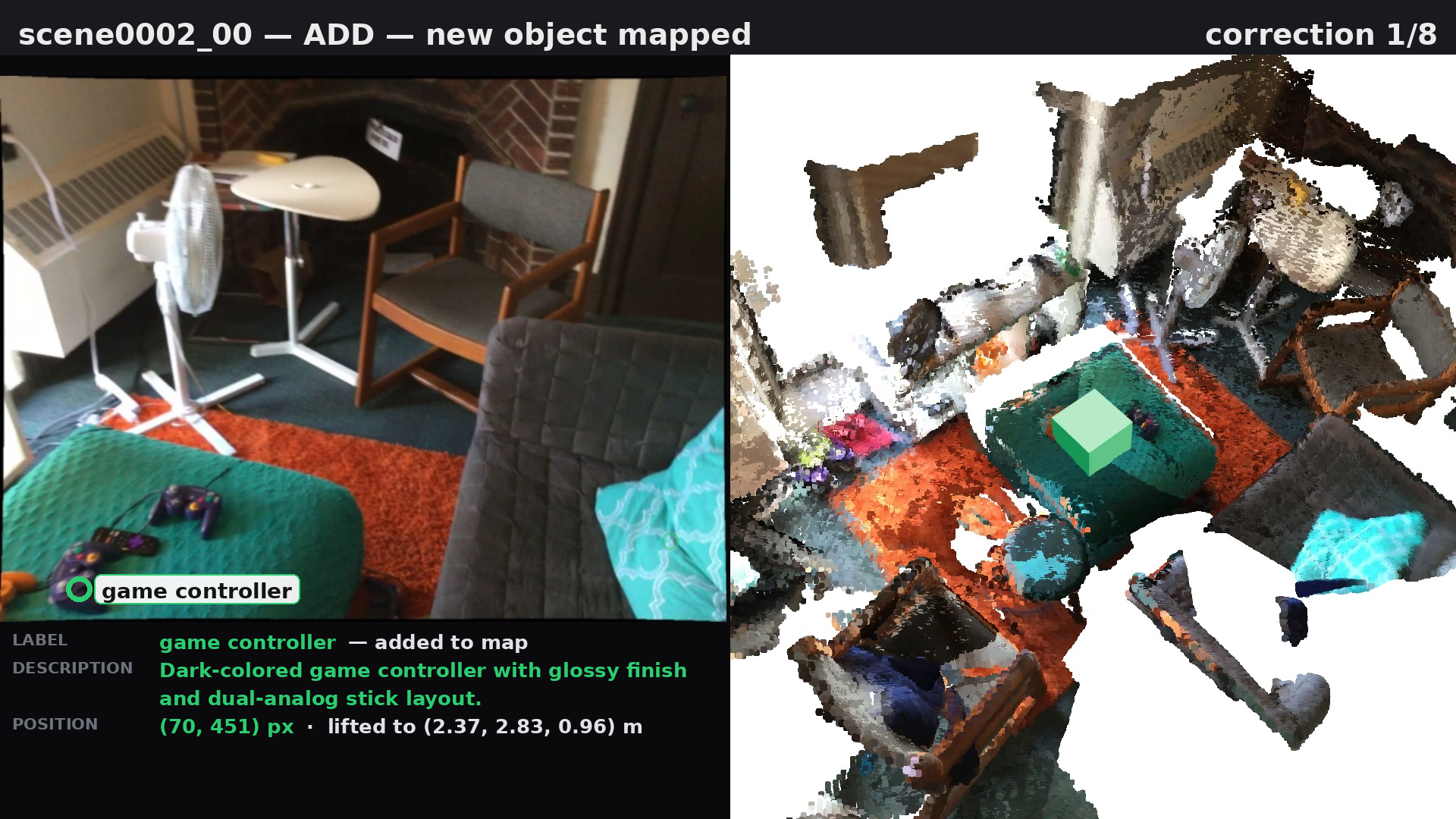}
        \caption{\add}
        \label{fig:edit:addition}
    \end{subfigure}
    \hfill
    \begin{subfigure}[t]{0.32\textwidth}
        \centering
        \includegraphics[width=\linewidth]{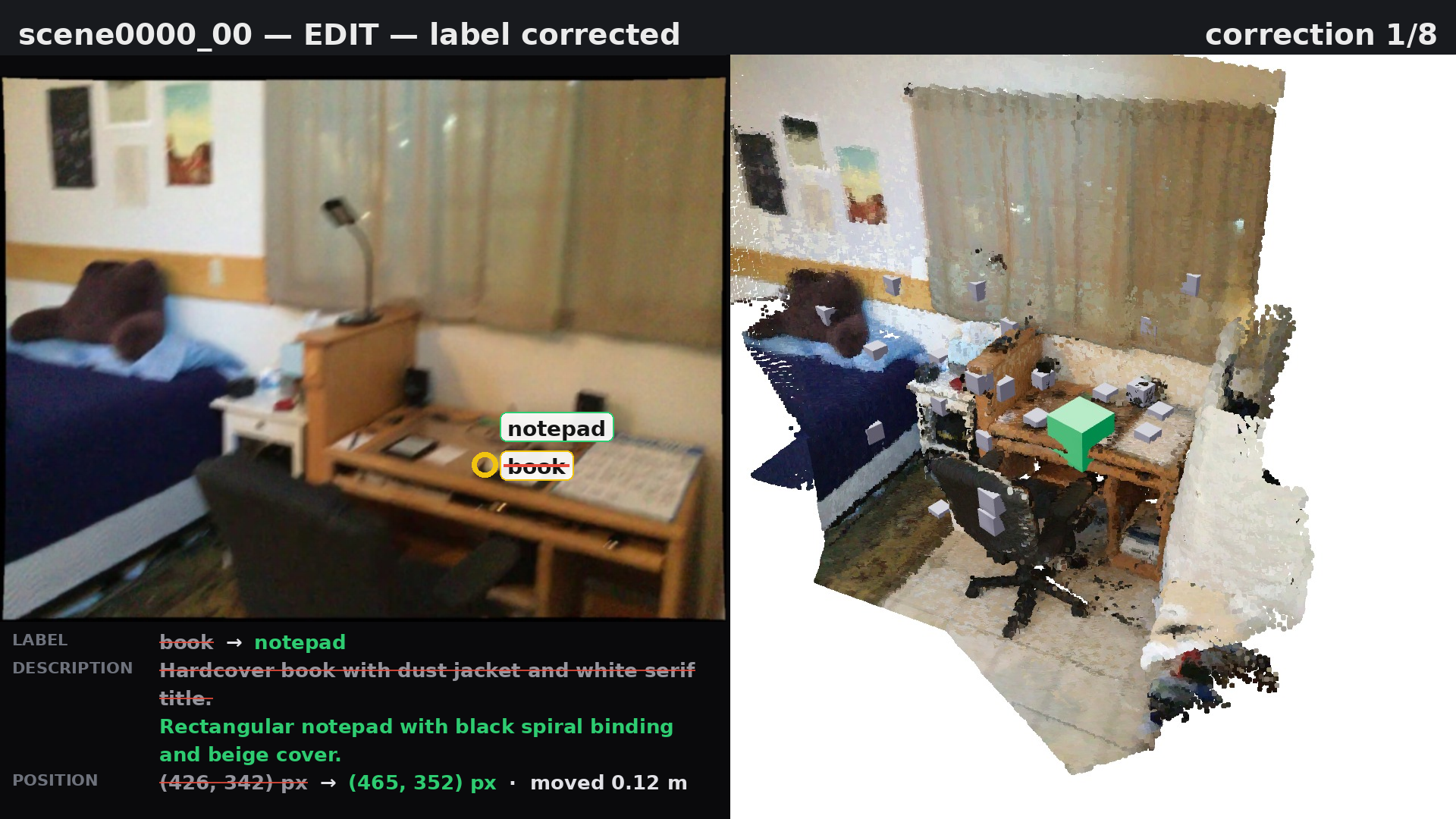}
        \caption{\edit}
        \label{fig:edit:relabel}
    \end{subfigure}
    \hfill
    \begin{subfigure}[t]{0.32\textwidth}
        \centering
        \includegraphics[width=\linewidth]{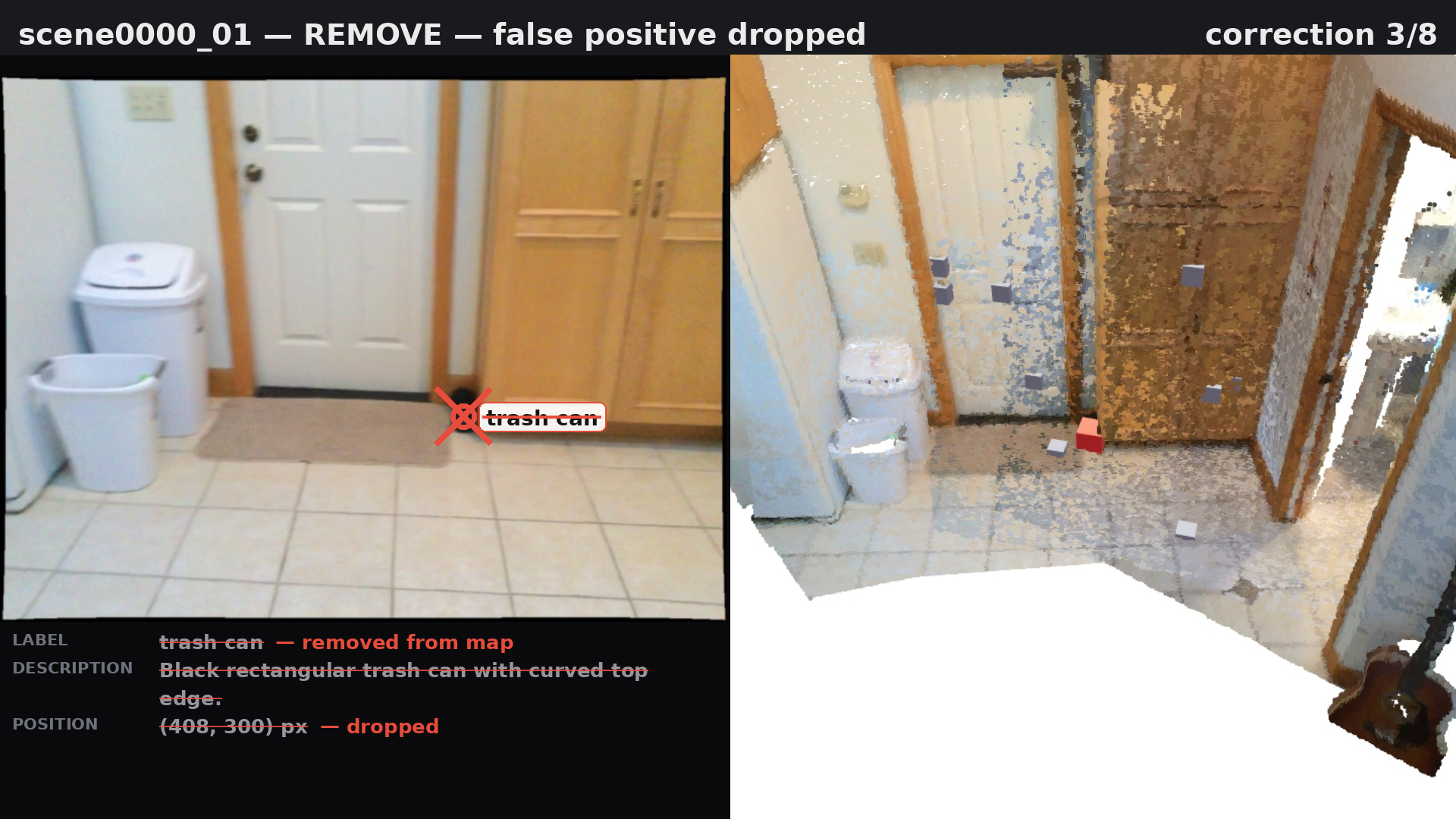}
        \caption{\remove}
        \label{fig:edit:removal}
    \end{subfigure}

    \caption{
    \textbf{Qualitative map-maintenance examples on a held-out test scene.}
    \textbf{(a) \add:} \modelname recognizes a game controller and adds it with an open-vocabulary label and description.
    \textbf{(b) \edit:} a closer view refines an existing entry from \emph{book} to \emph{notepad}.
    \textbf{(c) \remove:} additional visual evidence reveals a false-positive trash can, which is removed from the map.
    Colored point clouds are shown only to provide 3D context and are obtained by aggregating the posed RGB-D observations.
}
    \label{fig:qualitative_edits}
\end{figure*}

\section{\texorpdfstring{\modelname}{SceneLM}: model and training}
\label{sec:model}
We describe the model architecture and map update mechanism in \cref{sec:method-model}, the supervision task design in \cref{sec:method-tasks}, and present the training setup in \cref{sec:model-training}.

\subsection{Scene-map maintenance}
\label{sec:method-model}

\modelname represents the persistent scene map as object records
$M_t=\{o_i\}_{i=1}^N$, each containing a category label, short description, and 3D world position.
Map maintenance is modeled autoregressively as
$M_t=f_\theta(M_{t-1}, I_t)$,
where $I_t$ is the incoming posed RGB-D observation.

At each timestep, we select the objects in $M_{t-1}$ visible in $I_t$ using camera pose and depth-based visibility filtering, serialize them as structured text, and provide them together with the RGB image to the model.
\modelname predicts structured \add, \edit, and \remove operations that insert new objects, update existing labels, descriptions, or positions, and remove incorrect or redundant entries.
These operations jointly perform cross-view association and scene-map reconciliation within a single forward pass.
Predicted objects are localized by a 2D anchor in normalized $[0,1000]$ image coordinates, which is unprojected to 3D using depth and camera pose.
After applying the predicted operations, the updated map $M_t$ becomes the persistent state for the next timestep.
No external tracking, optimization, or geometric data association is used.

\subsection{Training tasks}
\label{sec:method-tasks}
The goal of training is to teach the model to maintain a scene map by predicting the updates required after each new observation, see \cref{fig:qualitative_edits}.
\pipelinename provides object annotations for individual images, but not the scene-map transitions required for this task.
We therefore algorithmically construct training examples that mimic required map updates from the pseudo-labels.
Each example consists of an input scene-map context, an image, and the corresponding set of update operations.
Training is organized into two complementary task families.
\add\ tasks teach the model to extend the scene map by inserting new objects, while \edit-\remove\ tasks teach it to correct an existing map through sparse update operations.

\textbf{\add\ tasks}
supervise the addition of new objects into the scene map.
We instantiate this supervision task through different sampling strategies that control the amount of prior information available:
In the \emph{no-context} setting, the model predicts all visible objects from an empty context and learns initial map construction.
In the \emph{partial-context} setting, a subset of objects is provided as an existing map, and the model must add only the missing objects without duplicating or reintroducing known ones.
Partial context is generated either through spatial cropping or random subsampling of objects, encouraging robustness to incomplete map states and cross-view association.
The model receives an image together with a partial or empty scene-map context and is trained to predict all remaining objects that should be inserted to get a complete map.

\textbf{\edit-\remove\ tasks}
supervise correction of an existing scene map.
The model receives a corrupted scene map together with the corresponding image and predicts a sparse set of \edit\ and \remove\ operations required to restore the correct map.
Objects not mentioned in the output are implicitly kept unchanged, yielding a sparse update script that may also be empty.
Corruptions include incorrect labels, perturbed object locations, duplicated entries, and hallucinated objects, and may appear individually or in combination.
A key design choice is how to generate corruption signals.
Sampling a single corruption rate per image leads to strong image-level correlations, encouraging degenerate policies that either modify most objects or none at all.
Instead, we corrupt each object independently, removing this global coupling and forcing object-wise decisions grounded in visual evidence.
This makes partial corrections the dominant training signal, aligning the learning objective with iterative scene-map maintenance.

\subsection{Training details}
\label{sec:model-training}
In all experiments we instantiate \modelname using 2B or 4B Qwen3-VL~\citep{qwen3vl2025}.
We train the vision encoder, projector, and language model jointly. 
Training runs for 3 epochs with AdamW, a cosine schedule, peak learning rate $2\times10^{-5}$, effective batch size 128, and max sequence length 4096, on 8 NVIDIA~A100 80GB GPUs. 
Full configuration is in \cref{tab:training-config}. 

\begin{figure}
    \centering
    \includegraphics[width=\linewidth]{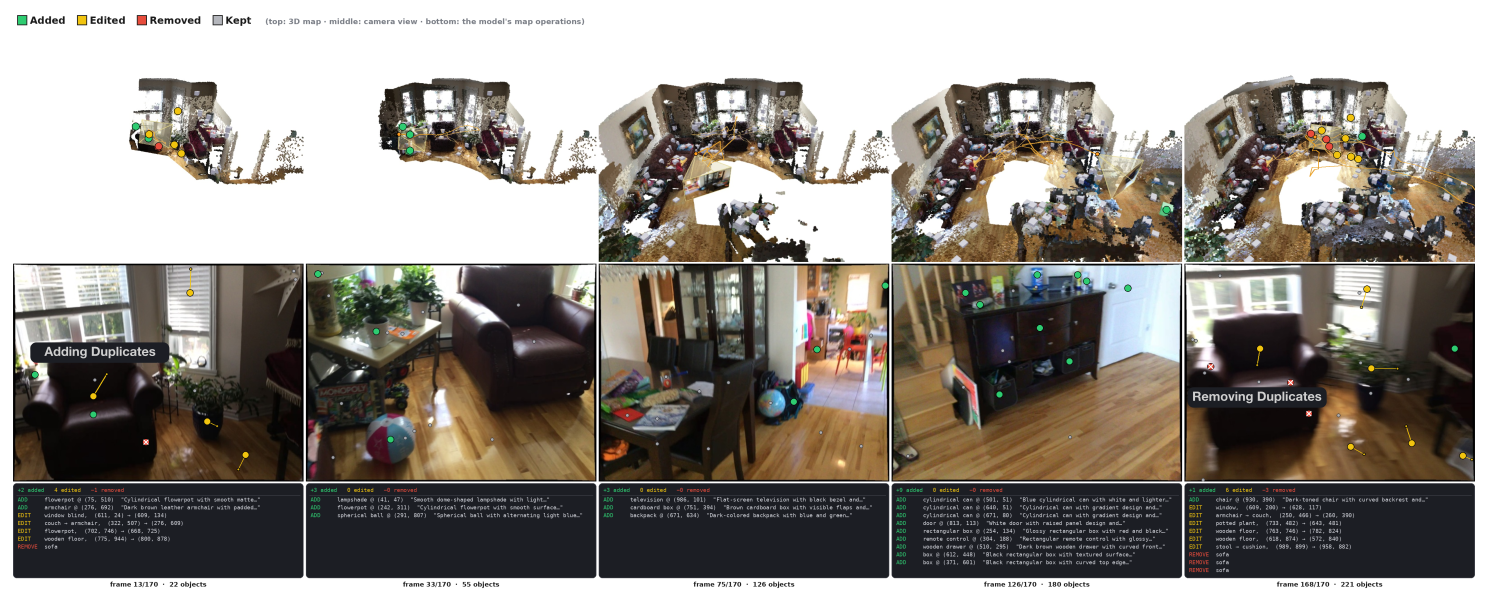}
    \caption{
\textbf{Scene-map evolution over a complete trajectory.}
Each column shows the accumulated 3D context (top), current RGB observation (middle), and predicted \add, \edit, and \remove operations (bottom).
\modelname incrementally maps diverse objects with open-vocabulary labels and descriptions, while later observations can reconcile earlier errors; here, duplicate objects are first added and subsequently removed.
}
    \label{fig:failure2}
\end{figure}

\begin{figure}
    \centering
    \includegraphics[width=\linewidth]{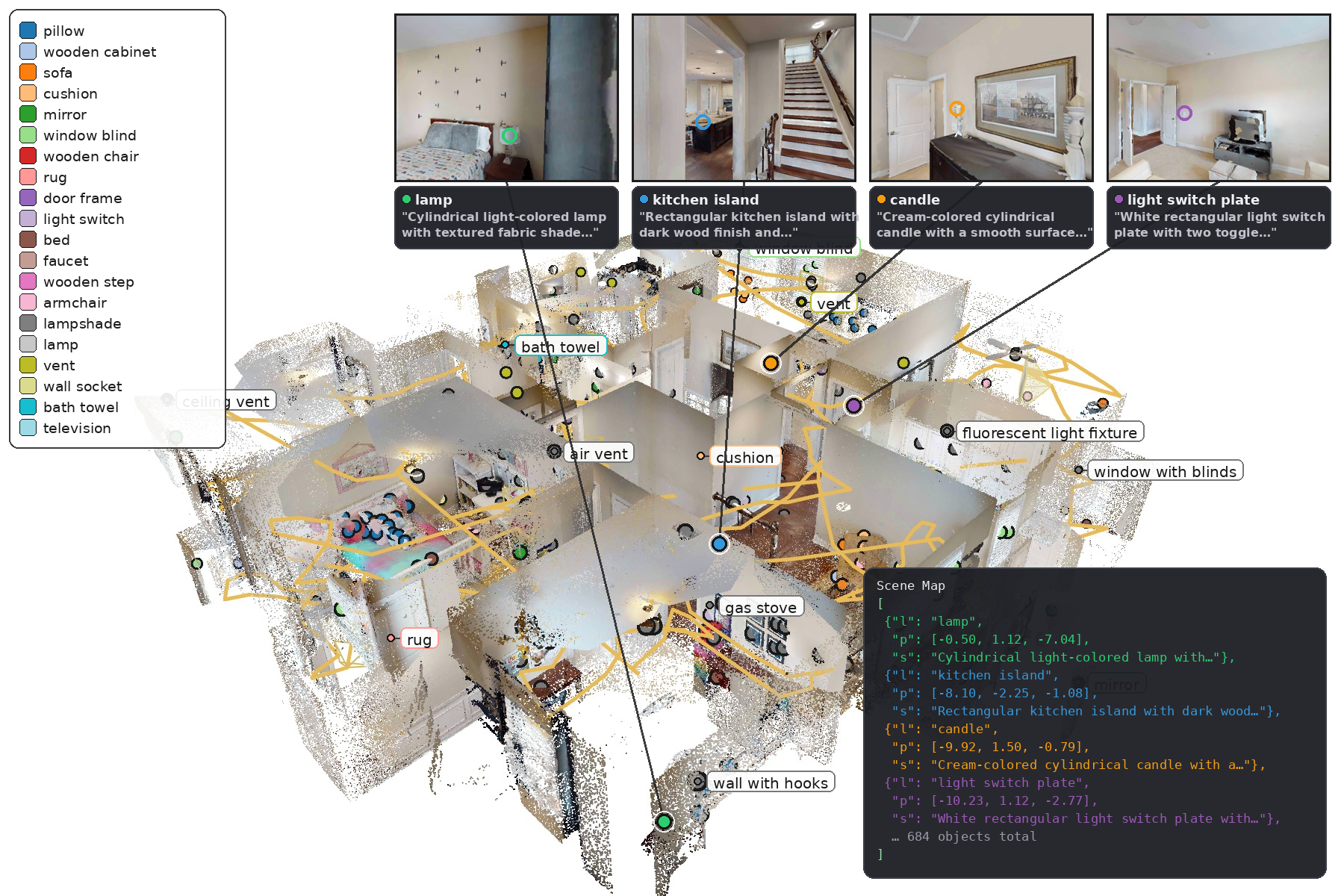}
    \caption{
    \textbf{Example scene map produced by \modelname on HM3D.}
    Colored markers show mapped object locations, with selected objects linked to the RGB observations in which they were identified.
    The inset shows the text-only scene memory, where each object is stored as a 3D position, noun phrase, and short description.
    }
    \label{fig:map1}
\end{figure}

\section{Experiments}
\label{sec:experiments}
Our experiments address four questions.
First, \emph{can a vision--language model learn the basic operations required to maintain a scene map?}
To answer this, we evaluate \add, \edit, and \remove in controlled single-frame experiments, comparing the finetuned model against off-the-shelf VLMs in \cref{sec:single-frame-performance}.
Second, in \cref{sec:exp-farm,sec:exp-tagmap}, \emph{can a single vision--language model maintain a competitive persistent open-vocabulary scene map using only a text-based representation as memory?}
Here we evaluate the resulting maps on two complementary open-vocabulary mapping benchmarks: language-grounded object retrieval following FARM~\citep{he2026farm}, and 3D object localization for navigation following TagMap~\citep{zhang2024tagmap} .
Qualitative examples of the mapping process and resulting scene maps are shown in \cref{fig:failure2,fig:map1}.
Third, \emph{which components are responsible for successful persistent mapping?}
To this end, \cref{sec:exp-analyses} synthesizes controlled ablations of proposal coverage, short object descriptions, quality filtering, and explicit \edit-\remove training.
Finally, \emph{what are the scene-memory and inference-time trade-offs of replacing a modular mapping backend with a single vision--language model?}
We compare scene-memory size and inference cost against existing systems and evaluate whether \modelname can operate online on embedded hardware through deployment on a quadruped in \cref{sec:efficiency,sec:real-deployment}.

\subsection{Setup}
\label{sec:exp-setup}
\parsection{Training Tasks}
We evaluate whether the model has learned the three core map-maintenance primitives: \add, \edit, and \remove.
For \add, the model receives an image with an empty scene map and predicts all visible objects.
Because the dataset vocabularies are incomplete, we report recall based on whether the predicted anchor lies inside the ground-truth object ($R_{\text{pos}}$), and whether both anchor and label match ($R_{\text{obj}}$).
These metrics provide a lower bound since spelling variants and synonyms are not resolved.
For \edit/\remove, the model receives a corrupted scene map and image and must keep, edit, or remove each object.
We evaluate whether the model identifies objects requiring intervention, selects the correct action, and improves corrupted labels or positions.
Full metric definitions and baseline details are provided in \cref{app:sec:per-frame-experiment}.

\parsection{Retrieval}
We follow the FARM~\citep{he2026farm} benchmark and baseline setup, using the 30 largest ReferIt3D~\citep{achlioptas2020referit_3d} scenes on ScanNet~\citep{dai2017scannet} and IRef-VLA~\citep{zhang2025iref} on HM3D~\citep{ramakrishnan2021hm3d}, and use FARM's implementations and evaluation setup for the mapping baselines.
Given a language query, we use the FARM retrieval system, without its visual embedding, to rank objects in the scene map produced by \modelname.
We report top-1 retrieval accuracy (Acc@1), Recall@5, Recall@10, and mean reciprocal rank (MRR), which rewards placing the first correct retrieval high in the ranking.
Because \modelname represents each object by a 3D point, label, and short description rather than an extent, we replace FARM's IoU-based matching criterion ($\tau=0.1$) with an extent-free evaluation in which a retrieval is correct if the predicted point lies inside the ground-truth 3D bounding box.
We compare against BBQ~\citep{linok2025beyond}, DAAAM~\citep{gorlo2025daaam}, FARM~\citep{he2026farm}, and RynnBrain~\citep{damo2026rynnbrain}, a frame-based 30B VLM that localizes queried objects directly from images without building a persistent map, under this point-in-box criterion.
Results under the original FARM protocol are provided in \cref{app:sec:farm-more-experiments}.

\parsection{Localization}
We follow the TagMap~\citep{zhang2024tagmap} benchmark on Matterport3D~\citep{chang2017matterport3d}.
Given a category query, the system retrieves the best matching object and selects the localization node closest to the object.
Following prior work, we report proposal-to-entity precision and entity-to-proposal recall at navigation thresholds $\tau\in\{0.1,0.5,1.0,2.0\}$\,m, macro-averaged over the 21 Habitat ObjectNav categories. 
Note that for ScanNet, HM3D, and Matterport3D, all evaluation scenes are excluded from both OVAL pseudo-label generation and \modelname training.

\subsection{Training Tasks Results}
\label{sec:single-frame-performance}

\parsection{\add}
Results are shown on the left of \cref{tab:single-frame}.
\modelname improves single-frame object prediction and removes the looping behavior observed in off-the-shelf VLMs, which repeat objects on up to 40\% of images.
Adding short descriptions or jointly training \edit-\remove leaves \add recall largely unchanged, while coarse coordinate quantization has little effect.
Random object ordering slightly reduces recall and can reintroduce looping.
\dense proposals increase recall by producing substantially more object candidates, whereas \sparse yields a more compact scene representation.

\parsection{\edit-\remove}
Results are shown on the right of \cref{tab:single-frame}.
Off-the-shelf VLMs perform poorly at the decision layer: they modify too many correct objects, yielding false-touch rates of 83--91\%, and achieve decision F1 scores around 30\%.
They also almost never perform \remove correctly, with removal recall approximately zero.
After finetuning, \modelname improves decision F1 to above 60\%, false touches fall to 10--25\%, and both \edit and \remove recall increase substantially.
Among correctly action-typed edits, predicted labels are corrected in 81--87\% of cases and predicted positions improve over the corrupted locations in 77--78\%.
These results show that the proposed finetuning changes both \emph{when} the model intervenes and \emph{which} corrective action it selects.

Taken together, these controlled experiments show that a vision--language model can learn the three basic operations required for scene-map maintenance: adding new objects and deciding when existing entries should be edited or removed.

\begin{table*}[t]
\centering
\scriptsize
\setlength{\tabcolsep}{1.5pt}
\renewcommand{\arraystretch}{0.6}

\begin{tabular}{
l
cccccccc
@{\hspace{8pt}}
cccccccc
}
\toprule

&
\multicolumn{8}{c}{\textbf{\add}}
&
\multicolumn{8}{c}{\textbf{\edit-\remove}}
\\

\cmidrule(lr){2-9}
\cmidrule(lr){10-17}

&
\multicolumn{2}{c}{COCO}
&
\multicolumn{2}{c}{LVIS}
&
\multicolumn{2}{c}{nuImages}
&
\multicolumn{2}{c}{Average}
&
\multicolumn{3}{c}{Decision}
&

&
\multicolumn{2}{c}{Action}
&
\multicolumn{2}{c}{Quality}
\\

\cmidrule(lr){2-3}
\cmidrule(lr){4-5}
\cmidrule(lr){6-7}
\cmidrule(lr){8-9}
\cmidrule(lr){10-12}
\cmidrule(lr){14-15}
\cmidrule(lr){16-17}

Model
& $R_{\text{pos}}\uparrow$ & $R_{\text{obj}}\uparrow$
& $R_{\text{pos}}\uparrow$ & $R_{\text{obj}}\uparrow$
& $R_{\text{pos}}\uparrow$ & $R_{\text{obj}}\uparrow$
& Loop$\downarrow$ & Obj./Img.
& F1$\uparrow$ & P$\uparrow$ & R$\uparrow$
& FT$\downarrow$
& Edit$\uparrow$ & Rem.$\uparrow$
& Label$\uparrow$ & Pos.$\uparrow$
\\

\midrule
\multicolumn{17}{l}{\textit{Ours and ablations}} \\

\sparse w Desc.
&65&41&42&21&37&27&0&8.7
&65&57&77&15&70&72&87&77\\

\sparse w/o Desc.
&65&41&42&21&37&27&0&9.0
&61&48&87&25&72&83&80&78\\

\dense w Desc.
&69&43&53&28&41&26&0&26.9
&66&64&69&10&71&57&87&78\\

\add-only
&65&40&42&20&38&25&0&8.5
&--&--&--&--&--&--&--&--\\

Random Order
&64&41&40&20&36&26&1&11.0
&63&51&84&22&68&78&82&78\\

Quantized-40
&67&42&43&21&38&27&0&8.6
&62&51&84&22&68&81&81&78\\

\midrule
\multicolumn{17}{l}{\textit{Off-the-shelf VLMs}} \\

Qwen3-VL-2B
&29&21&15&7&15&12&35&19.2
&27&19&49&52&50&0&19&3\\

Qwen3-VL-4B
&51&39&30&15&29&24&40&23.7
&29&18&82&91&85&0&21&2\\

Qwen3-VL-8B
&51&38&30&16&31&25&20&31.2
&28&18&76&86&81&0&27&4\\

RynnBrain-30B
&37&30&18&8&17&14&20&7.1
&31&19&80&83&82&0&17&7\\

\bottomrule
\end{tabular}

\caption{
Single-frame evaluation of \modelname-4b.
Finetuning eliminates looping and substantially improves scene-map correction, reducing false touches from 83--91\% to 10--25\% while enabling both \edit and \remove.
\dense proposals improve recall at the cost of more objects per image.
All metrics are percentages except \emph{Obj./Img.}; arrows indicate whether higher or lower is better.
}

\label{tab:single-frame}
\end{table*}

\subsection{Retrieval Results}
\label{sec:exp-farm}
We first evaluate whether the text-only persistent scene maps produced by \modelname support language-grounded object retrieval.
\modelname improves Recall@5, Recall@10, and mean reciprocal rank over FARM on both ScanNet and HM3D, while achieving similar or better top-1 retrieval accuracy.
Adding short object descriptions improves most retrieval metrics, suggesting that object-level semantic attributes provide useful information beyond the noun phrase alone.
The proposal strategy introduces a clear trade-off: \dense proposals increase object coverage but also redundancy, while \sparse proposals yield stronger overall retrieval on these benchmarks.
Removing \edit-\remove training causes the largest degradation, particularly on ScanNet, indicating that scene-state reconciliation is important for maintaining a consistent persistent map.
Removing quality filtering produces smaller but still measurable drops.
Overall, \modelname matches or exceeds prior mapping systems on top-1 accuracy and improves the ranking-based retrieval metrics, while storing only a text-based persistent scene representation.

\begin{table}[t]
\scriptsize
\centering
\renewcommand{\arraystretch}{0.6}
\begin{tabular}{lccccccccc}
\toprule
 & &\multicolumn{4}{c}{ScanNet} & \multicolumn{4}{c}{HM3D} \\
\cmidrule(lr){3-6} \cmidrule(lr){7-10}
Method & Venue & Acc@1 $\uparrow$ & R@5$\uparrow$ & R@10 $\uparrow$ & MRR $\uparrow$ & Acc@1 $\uparrow$ & R@5 $\uparrow$ & R@10 $\uparrow$ & MRR $\uparrow$ \\
\midrule
BBQ           & ICRA'25 & 22.6 & 34.3 & 34.9 & 27.6 & 4.2 & 8.6 & 10.1 & 6.1 \\
RynnBrain-30B &  & 38.9 & — & — & — & 4.3 & — & — & — \\
DAAAM         & CVPR'26  & 37.1 & — & — & — & 5.4 & — & — & — \\
FARM          &  & 32.0 & 63.0 & 75.0 & 45.7 & 6.3 & 17.2 & 25.1 & 11.7 \\
Qwen3-VL-2B $\dagger$ && 12.4 & 20.9 & 23.1 & 16.3 & 2.4 & 5.6 & 6.8 & 3.9 \\
Qwen3-VL-4B $\dagger$ && 25.6 & 44.7 & 48.5 & 34.1 & 4.0 & 10.1 & 13.6 & 6.8 \\
\bestrow\modelname-2B && 38.5 & 70.4 & 78.1 & 52.2 & 6.7 & 18.2 & 26.5 & 12.3 \\
\bestrow\modelname-4B && 39.3 & 71.3 & 79.2 & 52.6 & 6.6 & 18.8 & 26.5 & 12.5 \\
\midrule
\textbf{Ablations} && & & & & & & & \\
\multirow{4}{*}{\modelname-2B} &w/o Desc. & 36.7 & 70.3 & 78.3 & 51.0 & 6.0 & 17.3 & 24.8 & 11.5 \\
 &No Qual.  & 36.8 & 69.9 & 79.4 & 50.9 & 5.8 & 18.5 & 27.1 & 12.0 \\
 &\add-only & 29.7 & 54.6 & 61.3 & 40.7 & 6.5 & 17.6 & 24.2 & 11.8 \\
 &\dense    & 34.5 & 62.4 & 71.2 & 46.5 & 5.5 & 14.0 & 20.0 & 9.8 \\
\bottomrule
\end{tabular}
\caption{Scene-graph retrieval results and ablations. \modelname achieves strong point-on-box results on both ScanNet and HM3D. 
$\dagger$ \add only with ad-hoc rules reducing failure cases and looping behavior.}
\label{tab:farm-summary-point3dbox}
\end{table}

\subsection{Localization Results}
\label{sec:exp-tagmap}
We next evaluate whether the resulting maps support category-based object localization for navigation.
\cref{tab:tagmap-test18-objectnav21-macro} reports Matterport3D localization performance under the TagMap benchmark.
We report precision (P) and recall (R) at navigation distance thresholds of $0.1$, $0.5$, $1.0$, and $2.0$\,m.
\modelname improves precision over prior mapping systems at all thresholds, with the largest gains at $0.1$ and $0.5$\,m.
The proposal strategy exposes a precision--recall trade-off: \dense proposals increase recall through higher object coverage, while \sparse proposals yield the highest precision.
Removing \edit-\remove training reduces both precision and recall across nearly all thresholds, while removing quality filtering causes smaller degradations.
Additional ablations are provided in \cref{app:sec:ablation-more-experiments}.
Taken together with the retrieval results in \cref{sec:exp-farm}, these complementary evaluations show that a single vision--language model can maintain a competitive persistent open-vocabulary scene map using only text-based memory, supporting both language-grounded object retrieval and category-based localization for navigation.

\begin{table}[t]
\scriptsize
\centering
\renewcommand{\arraystretch}{0.6}
\begin{tabular}{lccccccccc}
\toprule
Method & Venue &P@0.1 $\uparrow$ & P@0.5 $\uparrow$ & P@1.0 $\uparrow$ & P@2.0 $\uparrow$ & R@0.1 $\uparrow$ & R@0.5 $\uparrow$ & R@1.0 $\uparrow$ & R@2.0 $\uparrow$ \\
\midrule
OpenScene & CVPR`23 & 31.0 & 36.0 & 41.0 & 46.0 & 12.0 & 29.0 & 61.0 & 88.0 \\
OpenMask3D & NeurIPS`23  & 29.0 & 35.0 & 41.0 & 48.0 & 5.0 & 19.0 & 41.0 & 60.0 \\
TagMap &CoRL`24 & 28.0 & 39.0 & 46.0 & 53.0 & 13.0 & 42.0 & 65.0 & 82.0 \\
\bestrow\modelname-2B &  & 41.4 & 48.8 & 54.4 & 60.8 & 28.0 & 63.9 & 71.6 & 78.4 \\
\bestrow\modelname-4B  &  & 43.2 & 50.0 & 55.5 & 62.1 & 28.7 & 66.4 & 76.6 & 81.7 \\
\bottomrule
\end{tabular}
\caption{
Matterport3D object localization under the TagMap benchmark.
P and R denote precision and recall at the indicated navigation distance thresholds in meters.
\modelname improves precision over prior mapping systems across all thresholds.
}
\label{tab:tagmap-test18-objectnav21-macro}
\end{table}

\subsection{Components Enabling Scene Mapping}
\label{sec:exp-analyses}
We next synthesize the ablations from the training-task, retrieval, and localization evaluations in \cref{sec:single-frame-performance,sec:exp-farm,sec:exp-tagmap} to identify which design choices enable persistent scene mapping.

\parsection{Coverage over precision}
The proposal source sets the system's coverage--precision operating point.
\dense proposals increase object coverage and localization recall, whereas \sparse proposals yield more compact maps and higher precision.
For persistent mapping, greater coverage is particularly valuable when missed objects are not observed again, whereas spurious entries can potentially be corrected through subsequent reconciliation.

\parsection{Textual object representations}
Short descriptions have little effect on single-frame localization but improve retrieval, indicating that they contribute primarily to object identity rather than detection.
The additional semantic detail helps distinguish objects with similar noun phrases during language-based retrieval.
At the same time, a text-only representation necessarily discards fine-grained visual information, which can introduce ambiguity when multiple instances share similar labels or descriptions.
The improvement from adding short descriptions suggests that richer textual attributes partially mitigate this ambiguity, but text remains a lossy representation and may not be optimal for all downstream tasks.

\parsection{Learning \edit and \remove}
Removing \edit-\remove training leaves single-frame \add performance largely unchanged but degrades both retrieval and localization.
These results show that maintaining a scene map as new observations arrive requires more than detecting and adding objects.
The model must also learn when to edit existing entries and remove stale or duplicate ones.

\parsection{Quality filtering and serialization choices}
Object ordering, coordinate quantization, and quality filtering have smaller effects than proposal coverage, semantic descriptions, and \edit-\remove training.
Removing quality filtering causes smaller but measurable degradations, indicating that it contributes to mapping performance without being its primary driver.

Overall, the proposal source and \edit-\remove training have the broadest effects on mapping performance, while short descriptions mainly help retrieval and quality filtering mainly improves precision.
Object ordering and coordinate quantization have only minor effects.
Additional ablations are provided in \cref{app:sec:ablation-more-experiments}.

\subsection{Efficiency}
\label{sec:efficiency} 
We evaluate scene-memory size and per-frame mapping latency on ScanNet and HM3D using an A100.
As shown in \cref{tab:efficiency}, \modelname uses approximately $6$--$12\times$ less scene memory than FARM: 3.5\,MiB vs.\ 23\,MiB on ScanNet and 10.8\,MiB vs.\ 125\,MiB on HM3D.
This reduction comes from storing only object-level textual state rather than feature-rich multi-view representations.
The trade-off is higher latency.
\modelname requires approximately 900\,ms per frame, while FARM is about $3\times$ faster.
Quantization increases throughput to $3.5$ frames/s on an RTX 5090 (\cref{tab:edge-rate}).
While VLM inference remains fairly slow, improving model efficiency is largely orthogonal to our mapping formulation, and continued advances in VLM inference can be directly inherited by \modelname.

\begin{table}[t]
\centering
\begin{minipage}[t]{0.56\linewidth}
  \centering
  \scriptsize
  \renewcommand{\arraystretch}{0.6}
  \setlength{\tabcolsep}{4pt}
  \begin{tabular}{lcccc}
    \toprule
    & \multicolumn{2}{c}{\textbf{\modelname}} & \multicolumn{2}{c}{\textbf{FARM}} \\
    \cmidrule(lr){2-3} \cmidrule(lr){4-5}
    Metric & ScanNet & HM3D & ScanNet & HM3D \\
    \midrule
    Scene memory (MiB/scene) $\downarrow$ & \textbf{3.5} & \textbf{10.8} & 23 & 125 \\
    Mapping time (ms/frame) $\downarrow$ & 965 & 843 & \textbf{275} & \textbf{258} \\
    \bottomrule
  \end{tabular}
\caption{
\textbf{Memory--latency trade-off on A100.}
\modelname uses $6$--$12\times$ less scene memory than FARM, but requires roughly $3\times$ longer per-frame mapping time.
}
  \label{tab:efficiency}
\end{minipage}\hfill
\begin{minipage}[t]{0.42\linewidth}
  \centering
  \scriptsize
  \renewcommand{\arraystretch}{0.6}
  \setlength{\tabcolsep}{4pt}
  \begin{tabular}{lcccc}
    \toprule
    & \multicolumn{2}{c}{\textbf{RTX 5090}} & \multicolumn{2}{c}{\textbf{Jetson Thor}} \\
    \cmidrule(lr){2-3} \cmidrule(lr){4-5}
    Precision & 4B & 2B & 4B & 2B \\
    \midrule
    BF16  & 1.36 & 2.59 & 0.21 & 0.53 \\
    FP8   & 1.91 & 3.09 & 0.33 & 0.83 \\
    NVFP4 & \textbf{2.36} & \textbf{3.54} & \textbf{0.51} & \textbf{0.86} \\
    \bottomrule
  \end{tabular}
  \vspace{-4pt}
\caption{
\textbf{Mapping throughput under quantization.}
Lower-precision inference increases throughput on both RTX 5090 and Jetson Thor, with NVFP4 giving the highest frame rate.
}
  \label{tab:edge-rate}
\end{minipage}
\vspace{-0.2em}
\end{table}

\subsection{Deployment on a quadruped}
\label{sec:real-deployment}

\begin{figure}
    \centering
    \includegraphics[width=\linewidth, trim={0 300pt 0 0},clip]{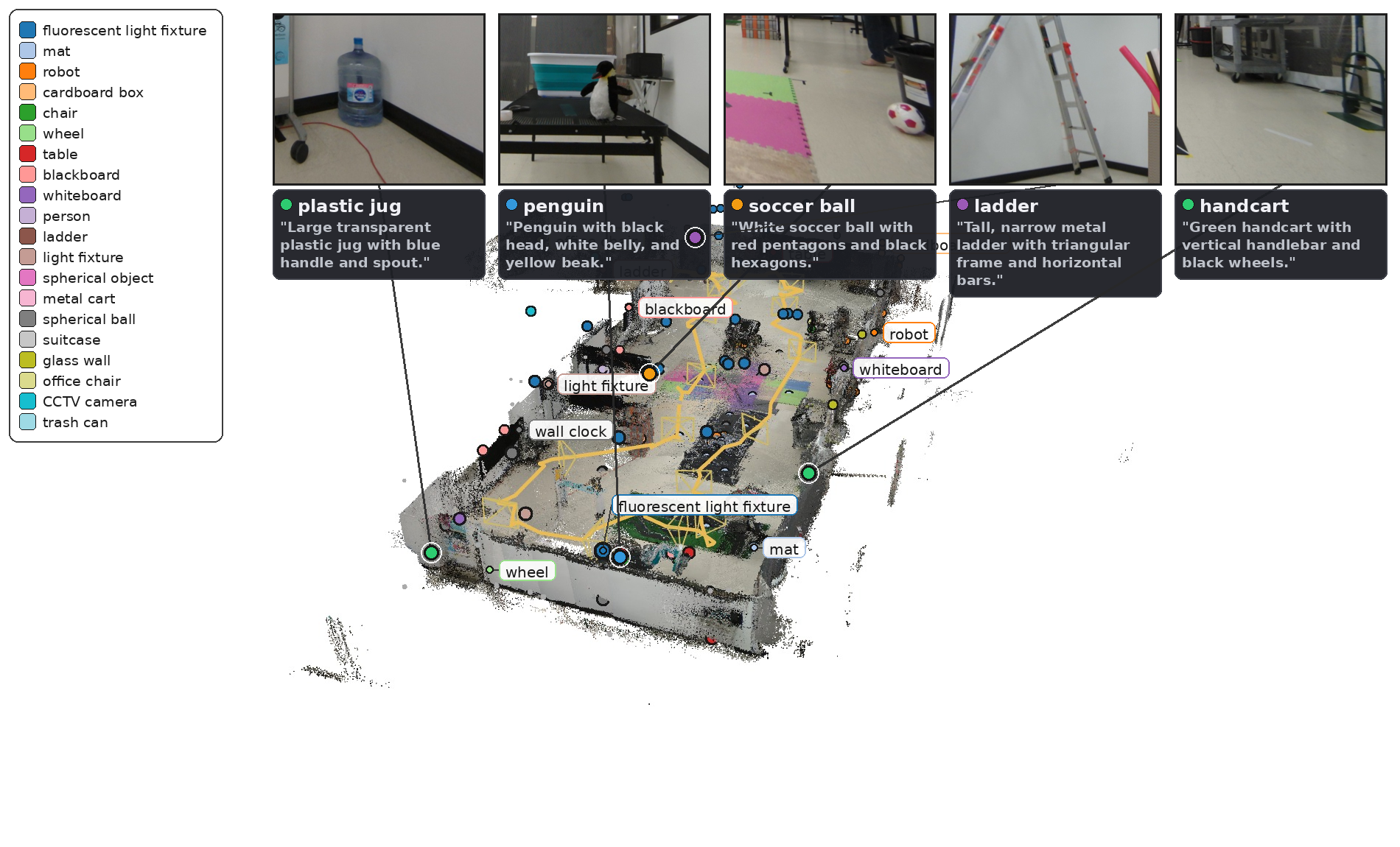}
    \caption{
\textbf{Real-world deployment of \modelname-2B-NVFP4 on a Spot quadruped.}
Selected mapped objects are shown within the resulting scene map from an unseen environment.
\modelname runs onboard an NVIDIA Jetson and updates the textual map online from posed RGB-D observations.
}
    \label{fig:deploy-map}
\end{figure}

We deploy \modelname on a Boston Dynamics Spot quadruped~\citep{bostondynamics2020spot} equipped with an NVIDIA Jetson and an Odin1 sensor~\citep{manifold2026odin1} providing posed RGB-D observations.
We evaluate three previously unseen environments: two indoor scenes and one outdoor scene.
Using \modelname-2B in NVFP4, we process a frame only after the robot has moved at least $0.5\,\mathrm{m}$ from the previous processed pose.
This keeps inference in a streaming regime without accumulating a frame backlog on the onboard hardware.
Across the three environments, 35 of 44 manually selected target objects are represented in the final maps.
These experiments show that the learned mapping behavior transfers beyond benchmark data and remains usable under real sensor noise, viewpoint changes, and scene clutter.
Taken together with the efficiency results in \cref{sec:efficiency}, they characterize the system-level trade-off: \modelname substantially reduces persistent scene-memory requirements through its text-only representation, whereas autoregressive VLM inference incurs high per-frame latency.
Nevertheless, onboard operation remains feasible when inference is coupled with motion-based frame selection rather than processing every incoming frame.
An example map is shown in \cref{fig:deploy-map}, and additional implementation details and qualitative results are provided in \cref{app:sec:deployment}.

\section{Conclusions}
\label{sec:conclusion}

We introduced \modelname, a scene--language model that maintains a persistent 3D scene map as structured text from posed RGB-D observations.
The scene memory consists only of object locations, labels, and short descriptions, without persistent visual embeddings or feature maps.
Across retrieval and localization benchmarks, \modelname matches or exceeds several modular mapping systems while using a substantially more compact scene representation.
Our ablations show that explicit \edit and \remove supervision is important for maintaining scene consistency over time, and \pipelinename provides the automatic annotations needed to train these behaviors.
The main limitation is inference cost, since each processed frame requires a full auto-regressive forward pass.
Nevertheless, quantized \modelname-2B can operate online on a quadruped using motion-based frame selection.
These results support a broader direction in which increasingly capable vision--language models absorb scene-map maintenance operations that are currently implemented by specialized mapping components.
Future work should improve inference efficiency and extend persistent scene modeling to longer-term and dynamic environments.
More broadly, persistent scene-state maintenance may become a useful capability of future vision--language models beyond robotic mapping.


\subsection*{Acknowledgements}
This work was supported in part by the DARPA TIAMAT program and by Mitsui O.S.K. Lines, Ltd. (MOL) through the Stanford Sustainable Mobility Center.
This work was also partially supported by the Wallenberg AI, Autonomous Systems and Software Program (WASP) funded by the Knut and Alice Wallenberg Foundation.
Computational resources were provided by NAISS at \href{https://www.nsc.liu.se/}{NSC Berzelius} and \href{https://www.c3se.chalmers.se/about/Alvis/}{C3SE Alvis}, partially funded by the Swedish Research Council, grant agreement no. 2022-06725.

\subsection*{AI use statement}

Pretrained vision--language models are used as components of the \pipelinename pseudo-annotation pipeline. 
They generate textual object descriptions and pseudo-annotations from real images and perform automated quality filtering. 
These outputs constitute machine-generated training data used to construct supervision for \modelname, but the input images themselves were not synthetically generated. 

We also used generative AI assistants, including coding agents, to assist with implementing training and experimental code and with parsing datasets. 
In addition, we used these tools to edit author-written text for clarity and readability and to assist in creating and revising scientific figures and visualizations.

We did not use generative AI tools to develop the scientific methodology or conceptual framework, design experiments, formulate mathematical claims, propose or refine hypotheses, translate text, perform qualitative or thematic data analysis, or interpret experimental results. 
Proof-related uses are not applicable to this empirical work.

All AI-assisted manuscript text, figures, code, and data-processing outputs were reviewed by the authors. 
AI-assisted code and dataset parsing were inspected and tested before being used to produce the reported results, and visualizations were checked against their underlying data. 
We take responsibility for the final content of this work, including all text, claims, code, and artifacts produced with the aid of generative AI.

\subsection*{Reproducibility statement}

The \pipelinename pseudo-annotation pipeline, its constituent models, and the resulting training data are described in \cref{sec:method-pipeline}.
The complete dataset mixture and the human evaluation of pseudo-label quality are provided in the appendix, and we are only using open-source datasets. 
The \modelname architecture, supervision tasks, and training procedure are described in \cref{sec:model}, with the full training configuration provided in \cref{sec:appendix-training}. 
The evaluation protocols and additional experimental details are provided in \cref{sec:exp-setup} and the corresponding appendix sections.

We, furthermore, release code for pseudo-annotation, dataset preparation, training, inference, and evaluation, together with environment specifications, configuration files, prompts, random seeds, and dataset splits. 
We additionally provide the model checkpoints used to obtain the reported results through Hugging Face. 
Because pseudo-annotation and model training contain stochastic components, independent reruns may produce slightly different outputs; the released configurations and checkpoint correspond to the results reported in this paper.


\bibliography{references}  
\bibliographystyle{iclr2027_conference}

\newpage
\appendix
\section{Appendix Content}

This appendix provides additional implementation details, experimental results, and qualitative examples that complement the main paper. It contains:
\begin{itemize}
    \item Details of the training data, including the dataset mixture (\cref{app:sec:dataset}) and an analysis of pseudo-label quality (\cref{app:sec:pseudo-label-quality}).
    \item The complete training configuration (\cref{sec:appendix-training}).
    \item Additional analysis of the per-frame mapping components, including prompt ablations for \add\ (\cref{app:sec:add-baselines}) and evaluation metrics for the \edit/\remove\ task (\cref{app:sec:edit-remove}).
    \item Additional experimental results and ablations for language-guided 3D retrieval (\cref{app:sec:more-experiments}).
    \item Additional qualitative mapping examples (\cref{sec:appendix-qualitative-examples}).
    \item Further details on deployment and edge inference (\cref{app:sec:deployment}).
\end{itemize}

\clearpage   
\section{Training data}
This appendix provides additional details on the training data used to train \modelname.
Since the pipeline operates on individual RGB images rather than videos or reconstructed scenes, it can readily incorporate data from a wide range of image sources.
We first summarize the datasets used for training in \cref{app:sec:dataset}, followed by an analysis of the pseudo-label quality in \cref{app:sec:pseudo-label-quality}.

\subsection{Datasets}
\label{app:sec:dataset}
\cref{app:tab:training-data} summarizes the training mixture used throughout the paper.
The selected datasets span indoor scans, embodied navigation, autonomous driving, and web imagery, providing broad visual diversity while requiring only single RGB images.
Since the pipeline is dataset-agnostic, additional image datasets can be incorporated without modifying the training procedure.
Note that we remove the evaluation scenes from HM3D, Matterport3D, and ScanNet from the training data, so that no evaluation scenes or frames entered \pipelinename or \modelname training, for a fair comparison.

\begin{table}[h]
  \centering
  \small
  \setlength{\tabcolsep}{4.5pt}
  \begin{tabular}{lrrrr}
    \toprule
    Dataset & Samples & Good obj.\ & Rejected obj.\ & Total obj.\ \\
    \midrule
    NaVILA~\cite{cheng2024navila}              & 433{,}718 & 3{,}146{,}945 & 745{,}264 & 3{,}892{,}209 \\
    HM3D~\cite{ramakrishnan2021hm3d}           & 227{,}248 & 1{,}361{,}945 & 351{,}291 & 1{,}713{,}236 \\
    COCO~2017~\cite{lin2014coco} (incl.\ unlab.)& 174{,}519 & 1{,}640{,}005 & 377{,}571 & 2{,}017{,}576 \\
    Matterport3D~\cite{chang2017matterport3d}  & 138{,}280 &   616{,}706   & 132{,}001 &   748{,}707   \\
    nuScenes~\cite{caesar2020nuscenes}         & 119{,}783 &   721{,}176   & 128{,}622 &   849{,}798   \\
    ScanNet~\cite{dai2017scannet}              &  95{,}150 &   667{,}574   & 154{,}102 &   821{,}676   \\
    nuImages~\cite{caesar2020nuscenes}         &  65{,}782 &   413{,}172   &  69{,}779 &   482{,}951   \\
    ODIN                                       &  10{,}560 &    91{,}946   &  23{,}388 &   115{,}334   \\
    SUN~RGB-D~\cite{song2015sunrgbd}           &   2{,}875 &    21{,}039   &   5{,}470 &    26{,}509   \\
    GrandTour (frontier subset)                &      385  &     1{,}314   &      262  &     1{,}576   \\
    \midrule
    \textbf{Total}                              & \textbf{1{,}268{,}300} & \textbf{8{,}681{,}822} & \textbf{1{,}987{,}750} & \textbf{10{,}669{,}572} \\
    \bottomrule
  \end{tabular}
  \caption{\textbf{Training mixture, per dataset} (epsilon pack, YOLOE
  proposals). Samples are RGB frames; objects are per-frame records emitted by the
  pipeline. ``Good'' objects (QC-passed) are ADD targets and carry the loss;
  ``rejected'' objects supply REMOVE/EDIT supervision. The COCO row folds in the
  unlabeled split.}
  \label{app:tab:training-data}
\end{table}

\subsection{Pseudo-label quality}
\label{app:sec:pseudo-label-quality}
The quality check decides \emph{which} proposals to keep; a separate question is
how accurate the kept labels are in absolute terms. 
We answer it with a \emph{golden set}: 100 images (30 COCO, 30 NaVILA, 25 nuScenes, 15 GrandTour) on
which a human judged every pseudo-labeled object correct or incorrect at its anchor point. 
Over $\sim$4{,}000 judged objects, \textbf{$71.8\%$ are correct} (Table~\ref{tab:golden}). 
Two patterns matter. 
First, quality tracks the proposal source exactly as the trained map does: \sparse proposals are $84.8\%$ correct against $68.6\%$ for \dense, a $+16$-point precision gap that prefigures the \sparse-precision / \dense-recall split on the cumulative map (Section~\ref{sec:exp-analyses}). 
Second, interior anchors are cleaner than center points ($71.8\%$ vs.\ $63.0\%$), confirming the anchor choice we make. 
Quality also varies by domain (nuScenes $82\%$, GrandTour $81\%$, COCO $72\%$, NaVILA $68\%$).
The takeaway is that the raw proposals are good but not clean, which is exactly why the quality gate, and the model's learned REMOVE behavior trained on what it rejects, are needed.

\begin{table}[h]
  \centering
  \small
  \begin{tabular}{lc}
    \toprule
    Slice & \% correct\\
    \midrule
    Overall (interior anchor)        & 71.8\\
    \quad \sparse proposals            & \textbf{84.8}\\
    \quad \dense proposals     & 68.6\\
    \bottomrule
  \end{tabular}
  \caption{\textbf{Pseudo-label accuracy} on the golden set (100 human-verified
  images). \% of pseudo-labeled
  objects judged correct, by proposal source.}
  \label{tab:golden}
\end{table}

\subsection{Quality Check}
\label{app:sec:quality-control}
The per-object keep/reject decision is made by a VLM \emph{quality check}, and its calibration matters: a stricter gate yields cleaner \add targets but more \edit-\remove fuel, and it must run over all $10.7$M objects, ruling out per-call API pricing.
We benchmark candidate quality checks against the human golden labels (\cref{tab:judge}). 
Frontier APIs are the most accurate (Gemini~3.1~Pro reaches MCC $0.64$) but carry a per-call cost we cannot pay at this scale. 
Our self-hosted Qwen3-VL-30B reaches MCC $0.47$ and balanced accuracy $0.74$---competitive with the
best affordable alternative---while agreeing with the human on $78\%$ of objects (Cohen's $\kappa=0.47$, moderate) and retaining a pool in which $84\%$ of kept objects are human-confirmed correct.

We ablate prompt-level design choices in \cref{tab:quality-check-ablation}, flipping one choice at a time off a base configuration. Three findings drive the deployed setup. 
(i)~Dropping the drawn point annotation and grounding the judge on the plain image maximizes recall on incorrect objects ($0.78$) but over-rejects, dropping agreement to $67\%$. 
(ii)~Reason-first prompting and explicit pixel grounding each curb this over-rejection, raising agreement to $\approx77\%$. 
(iii)~Majority voting gives no measurable gain over a single pass, and zooming on the object raises
agreement only by catching fewer bad objects. 
We therefore deploy Qwen3-VL-30B per object on the interior anchor, judging point and label from the plain image, with reason-first prompting, pixel grounding, no voting, and near-duplicate suppression.
Relative to the base configuration this combination lifts agreement from $73\%$ to $78\%$ and $\kappa$ from $0.39$ to $0.47$, trading peak bad-recall ($0.78\!\rightarrow\!0.63$) for a better-calibrated gate that does not needlessly inflate the removal pool.

The gate acts as a precision control rather than a recall booster: removing it slightly raises downstream recall but inflates predictions per image to $12.1$ (ground truth $7.25$) from the gated $10.4$.

\begin{table}[t]
  \centering
  \small
  \begin{tabular}{lccc}
    \toprule
    Judge & MCC & Bal.\ acc.\ & Recall-bad \\
    \midrule
    Gemini~3.1~Pro                       & 0.64 & 0.81 & 0.74 \\
    Grok~4.3                             & 0.48 & 0.74 & 0.68 \\
    Claude~Sonnet~4                      & 0.29 & 0.66 & 0.63 \\
    \textbf{Qwen3-VL-30B (used)}         & 0.47 & 0.74 & 0.63 \\
    \bottomrule
  \end{tabular}
  \caption{\textbf{Quality-control judge vs.\ human golden labels}, best
  configuration per model. MCC and balanced accuracy weight both classes;
  \emph{recall-bad} is the fraction of human-rejected objects the judge catches,
  the operative quality-gate metric. 
  \vspace{12pt} 
  }
  \label{tab:judge}
\end{table}

\begin{table}[t]
\centering
\begin{tabular}{lcccccc}
\toprule
Judge variant  & BalAcc & RecBad & Rec & $\kappa$ & Agree  \\
\midrule
Base: per-object, annotated, point+label, reason & 70.6\% & 63.9\% & 77.2\% & 0.39 & 73.2\%  \\
Query mode: batch (1 call/image) & 67.5\% & 60.6\% & 74.4\% & 0.33 & 70.2\%\\
Image: plain (no point annotation) & 70.1\% & 77.5\% & 62.8\% & 0.34 & 67.2\%  \\
Reasoning: none & 70.5\% & 64.7\% & 76.2\% & 0.39 & 72.8\%  \\
Reasoning: reason-first & 67.0\% & 42.5\% & 91.5\% & 0.38 & 76.6\%  \\
Abstention: allow skip & 65.2\% & 47.1\% & 83.4\% & 0.30 & 74.9\%  \\
Zoom-on-object: crop 0.25 & 68.5\% & 45.0\% & 92.0\% & 0.41 & 77.8\%  \\
Zoom-on-object: crop 0.50 & 70.8\% & 52.0\% & 89.7\% & 0.45 & 78.3\%  \\
Super-resolution & 72.3\% & 67.1\% & 77.5\% & 0.42 & 74.4\% \\
Pixel grounding & 68.9\% & 50.4\% & 87.3\% & 0.40 & 76.2\%  \\
\bottomrule
\end{tabular}
\caption{Controlled Quality Check prompt ablation on Qwen3-VL-30B: each row flips one design choice off the base configuration.}
\label{tab:quality-check-ablation}
\end{table}

\clearpage
\section{Training configuration}
\label{sec:appendix-training}

Objects are represented by a 2D anchor point in normalized $0$--$1000$ image coordinates, a noun-phrase label, and a short description.
Training examples are sampled from the pseudo-label and synonym/distractor pools of \cref{sec:method-pipeline}.
We sample \add\ and \edit-\remove\ tasks with probability $0.7$ and $0.3$, respectively.
Within \edit-\remove, $40\%$ of examples are no-ops with an empty target set, while the remaining examples are distributed evenly across attribute edits, removals, and mixed corruptions.
Within \add, half of the examples provide a partial scene map and require prediction of the remaining visible objects.

Rather than sampling a single corruption rate per image, we corrupt each object independently with probability $q\sim\mathcal{U}[0.05,0.20]$.
This removes image-level correlations and forces the model to decide independently which objects require modification, while naturally making single-object corrections common during training.

To avoid ambiguity between operations, perturbation magnitudes occupy disjoint ranges in anchor space: no-op jitter ($\approx1\%$ of the frame), edit displacements ($1.5$--$10\%$), and removal displacements ($\ge30\%$).
Finally, scene-map entries are serialized in spatial order and re-sorted after corruptions are inserted, preventing the model from exploiting list position to identify edited or hallucinated objects.

Table~\ref{tab:training-config} gives the full training configuration for the main \modelname runs.

\begin{table}[h]
  \centering
  \small
  \begin{tabular}{ll}
    \toprule
    Setting & Value \\
    \midrule
    Base model                            & Qwen3-VL-(4B/2B)-Instruct~\cite{qwen3vl2025} \\
    Trainable parameters                  & vision encoder $+$ projector $+$ LM (full fine-tune) \\
    Precision                             & bf16 \\
    Optimizer                             & AdamW \\
    Learning-rate schedule                & cosine decay, warmup ratio $0.05$ \\
    Peak learning rate                    & $2 \times 10^{-5}$ \\
    Weight decay                          & $0.05$ \\
    Epochs                                & $3$ \\
    Per-device train batch size           & $16$ \\
    Gradient accumulation steps           & $1$ \\
    Effective batch size                  & $128$ \\
    Sequence length (\texttt{model\_max\_length}) & $4096$ \\
    Image-token pixel bounds              & $\text{min}=784,\ \text{max}=50{,}176$ \\
    Distributed strategy                  & DeepSpeed ZeRO-2 \\
    Hardware                              & 8$\times$NVIDIA A100 (80~GB) per run \\
    Object-ordering supervision           & \texttt{shuffle\_object\_order=structured} \\
    Loss reduction                        & \texttt{token\_mean} \\
    Training-task mixture                 & ADD $+$ EDIT $+$ \texttt{refine} $+$ \texttt{dual\_crop} (Section~\ref{sec:method-tasks}) \\
    Serving                               & vLLM~\cite{kwon2023vllm} \\
    \bottomrule
  \end{tabular}
  \caption{\textbf{Training configuration} (\modelname headline run).}
  \label{tab:training-config}
\end{table}

\clearpage   

\newpage
\section{Per-frame component details}
\label{app:sec:per-frame-experiment}
This section provides additional details on the per-frame experiments presented in \cref{sec:single-frame-performance}.
We first analyze the prompts used for the single-frame \add\ baselines, followed by a detailed description of the evaluation protocol for the \edit/\remove\ task.

\subsection{\add - baselines}
\label{app:sec:add-baselines}
The performance of VLMs is highly dependent on the prompting strategy.
To ensure a fair comparison, we evaluate a diverse set of prompts for both Qwen3-VL-4B and Qwen3-VL-8B.
The resulting detection performance is summarized in \cref{tab:prompts-det-summary-val}, while the full prompt templates are provided in the accompanying code release.

The experiments demonstrate substantial variation across prompts.
Prompts encouraging exhaustive object discovery generally achieve higher localization and labeling accuracy, but also increase the likelihood of repetitive generation (loop behavior) and produce substantially more object predictions per image.
The prompt used throughout the paper provides the best trade-off between detection quality and stable generation.

\begin{table}[h]
\setlength{\tabcolsep}{3pt}
\renewcommand{\arraystretch}{0.75}
\centering
\begin{tabular}{llcccccccc}
\toprule
 & & \multicolumn{2}{c}{COCO (\%)} & \multicolumn{2}{c}{LVIS (\%)} & \multicolumn{2}{c}{nuImages (\%)} & \multicolumn{2}{c}{Average} \\
\cmidrule(lr){3-4}\cmidrule(lr){5-6}\cmidrule(lr){7-8}\cmidrule(lr){9-10}
Model & Prompt & loc & label  & loc& label & loc & label & Loop (\%) & Pred./img \\
\midrule
\multirow{10}{*}{Qwen3-VL-4B}& Exhaustive   & 51 & 39 & 30 & 15 & 29 & 24 & 40 & 23.67 \\
 & Few-shot     & 51 & 40 & 29 & 15 & 31 & 25 & 23 & 17.56 \\
 & LU           & 19 & 15 & 12 & 7  & 1  & 1  & 4  & 50.93 \\
 & LUD          & 4  & 3  & 2  & 1  & 0  & 0  & 1  & 47.80 \\
 & LUS          & 26 & 20 & 15 & 8  & 4  & 3  & 1  & 35.58 \\
 & LXY          & 45 & 35 & 25 & 13 & 20 & 16 & 21 & 11.94 \\
 & LXY3 System  & 44 & 35 & 22 & 11 & 26 & 22 & 26 & 9.44 \\
& Expert       & 47 & 37 & 26 & 13 & 28 & 24 & 34 & 12.01 \\
 & Systematic   & 49 & 39 & 26 & 14 & 30 & 25 & 28 & 13.69 \\
& Terse        & 50 & 40 & 28 & 15 & 26 & 21 & 32 & 27.16 \\
\midrule
\multirow{10}{*}{Qwen3-VL-8B} & LU           & 37 & 28 & 20 & 10 & 21 & 17 & 11 & 8.14 \\
 & LUD          & 35 & 27 & 18 & 9  & 21 & 17 & 12 & 7.49 \\
 & LUS          & 44 & 28 & 24 & 11 & 23 & 17 & 5  & 7.06 \\
 & LXY          & 40 & 30 & 24 & 13 & 20 & 16 & 23 & 17.44 \\
 & Exhaustive   & 42 & 29 & 26 & 13 & 24 & 19 & 36 & 35.09 \\
 & Few-shot     & 51 & 38 & 30 & 16 & 25 & 20 & 20 & 31.20 \\
 & LXY3 System  & 39 & 30 & 21 & 11 & 25 & 20 & 16 & 9.42 \\
 & Expert       & 45 & 32 & 28 & 14 & 26 & 21 & 37 & 28.28 \\
 & Systematic   & 35 & 24 & 22 & 11 & 19 & 15 & 29 & 22.62 \\
 & Terse        & 30 & 19 & 18 & 8  & 13 & 10 & 24 & 19.86 \\
\bottomrule
\end{tabular}
\caption{Prompt ablation for single-frame object prediction.}
\label{tab:prompts-det-summary-val}
\end{table}

\subsection{\edit/\remove}
\label{app:sec:edit-remove}
The \edit/\remove\ task evaluates whether the model can identify incorrect objects in the scene representation and determine the appropriate corrective action.
For completeness, \cref{app:tab:fix-task-metrics} defines all evaluation metrics used in the main paper.

The evaluation is organized into three stages.
First, the decision layer measures whether the model correctly determines if an object should be kept or modified.
Second, the action layer evaluates whether the model correctly predicts an \edit\ or \remove\ operation.
Finally, the quality layer measures the correctness of the predicted edits by evaluating both the corrected object label and the updated object position.

\begin{table}[h]
\centering
\small
\renewcommand{\arraystretch}{1.25}
\begin{tabularx}{\linewidth}{@{}llX@{}}
\toprule
Metric & Formula & Definition \\
\midrule
\multicolumn{3}{@{}l}{\textit{Decision layer} — for each object, keep vs.\ address (edit $\cup$ remove)} \\
\addlinespace[2pt]
Dec P (precision)        & $\mathrm{TP}/(\mathrm{TP}+\mathrm{FP})$ & Of the objects the model \emph{touched} (edited or removed), the fraction that genuinely needed fixing. Low $\Rightarrow$ the model meddles with objects it should have left alone. \\
\addlinespace[2pt]
Dec R (recall)           & $\mathrm{TP}/(\mathrm{TP}+\mathrm{FN})$ & Of the objects that \emph{needed} fixing, the fraction the model touched. Low $\Rightarrow$ the model misses corruptions and leaves them in the scene. \\
\addlinespace[2pt]
Dec F1                   & $2PR/(P+R)$ & Harmonic mean of decision precision and recall: the headline ``did it act on the right objects, and only those?''. The action-type and quality metrics are all conditioned on getting the decision right first. \\
\addlinespace[2pt]
FalseTouch ($\downarrow$) & $\mathrm{FP}/(\mathrm{FP}+\mathrm{TN})$ & Of the objects that should be \emph{left alone}, the fraction the model wrongly touched (lower is better). The clean-object error rate (denominator $=$ all keep objects), companion to decision precision. \\
\midrule
\multicolumn{3}{@{}l}{\textit{Action-type layer} — among addressed objects, edit vs.\ remove} \\
\addlinespace[2pt]
Edit R (recall)          & $\mathrm{TP}_{e}/(\mathrm{TP}_{e}+\mathrm{FN}_{e})$ & Of the objects whose correct action is \textsc{edit}, the fraction the model both addressed \emph{and} action-typed as edit (removing or missing it counts against). \\
\addlinespace[2pt]
Rem R (recall)           & $\mathrm{TP}_{r}/(\mathrm{TP}_{r}+\mathrm{FN}_{r})$ & Of the objects whose correct action is \textsc{remove}, the fraction the model both addressed \emph{and} action-typed as remove (hallucinations, duplicates, big shifts, \dots). \\
\midrule
\multicolumn{3}{@{}l}{\textit{Quality layer} — among objects correctly action-typed as edit} \\
\addlinespace[2pt]
Label acc                & $n^{\mathrm{ok}}_{\mathrm{label}}/n_{\mathrm{label}}$ & Among label-corrupted objects the model correctly chose to edit, the fraction whose predicted label matches the clean ground-truth label (synonym-aware). \\
\addlinespace[2pt]
Pos ok                   & $n_{\mathrm{closer}}/n_{\mathrm{pos}}$ & Among position-corrupted objects the model correctly chose to edit, the fraction whose predicted point landed \emph{closer} to the true location than the corrupted point was (residual after $<$ residual before); credits any improvement, not pixel-perfect placement. \\
\bottomrule
\end{tabularx}
\caption{Fix-tasks (\edit-\remove) evaluation metrics. The model is shown a context of objects (each with an \texttt{id}) in which some are deliberately corrupted, and must decide per object whether to \emph{keep} or \emph{address} it, where addressing is an \emph{edit} (fix the label and/or point) or a \emph{remove}. Scoring proceeds in three layers, each conditioned on the previous. The decision confusion matrix counts objects as \textbf{TP} (needed fixing and was touched), \textbf{FP} (touched but should have been kept --- a ``false touch''), \textbf{FN} (needed fixing but was kept), and \textbf{TN} (correctly left alone); $P,R$ in Dec~F1 are Dec~P and Dec~R. Subscripts $e,r$ denote the edit / remove action-type confusion counters. All metrics are micro-averaged over every object in the evaluation set; an undefined metric (zero denominator) is reported as ``---''.}
\label{app:tab:fix-task-metrics}
\end{table}

\clearpage
\section{Additional Experiments}
\label{app:sec:more-experiments}
This section reports additional ablation studies and supplementary experiments that did not fit in the main paper.
We first analyze several training design choices, followed by additional FARM-style experiments that estimate object extent during retrieval.

\subsection{Additional Ablations}
\label{app:sec:ablation-more-experiments}

\parsection{4B ablation}
We perform ablations for the 4b-\modelname when using or not using short descriptions during training and inference, \dense object proposal, \add-only, without the quality check.
Results are displayed in \cref{tab:farm-4b-ablations} and follow the conclusions drawn in the main paper on the 2b model.
On the TagMap benchmark, \cref{tab:tagmap-4b-ablations} show the ablations and importantly show that \dense-\modelname can be useful in certain scenarios.

\parsection{Loss reduction}
The language modeling loss is naturally accumulated per token, causing training examples with many objects or longer descriptions to contribute more to the optimization.
\Cref{tab:farm-loss-reduction} compares three reduction strategies: token mean, sample mean, and object mean.
The downstream retrieval performance is largely unchanged, indicating that the method is not particularly sensitive to this design choice.

\parsection{Dataset weighting}
The training mixture contains datasets of widely different sizes.
We therefore investigate three sampling strategies: the default sampling distribution, square-root weighting based on dataset size, and uniform dataset sampling with capped upsampling to avoid overfitting the smallest datasets.
As shown in \cref{tab:farm-data-weights}, uniform sampling improves performance on ScanNet but slightly reduces performance on HM3D, resulting in only minor differences overall.

\parsection{Model size}
We compare 2B, 4B, and 8B variants of \modelname\ in \cref{tab:farm-model-size}.
For practical reasons (fitting training on available GPU) the 8B model is trained without short descriptions.
While not perfectly comparable, the 2B model achieves the strongest overall retrieval performance despite having fewer parameters.
Increasing the model size does not improve downstream retrieval, suggesting that the bottleneck lies in the dataset, mapping formulation, and/or supervision rather than model capacity.

\parsection{Image resolution}
We evaluate the effect of the maximum image resolution used during both training and inference.
Internally, the limit is specified as a maximum number of visual tokens, but for readability we report the equivalent image resolution (e.g., resolution 316 corresponds to a maximum of 100\,352 pixels).
The results in \cref{tab:farm-img-res} show that increasing the resolution generally improves retrieval performance up to a point, after which the gains saturate.
The 2B model is more sensitive to this hyperparameter than the 4B model.

\begin{table}[h]
\centering
\begin{tabular}{lcccccccc}
\toprule
 & \multicolumn{4}{c}{ScanNet} & \multicolumn{4}{c}{HM3D} \\
\cmidrule(lr){2-5} \cmidrule(lr){6-9}
Method & Acc@1 & R@5 & R@10 & MRR & Acc@1 & R@5 & R@10 & MRR \\ \midrule
(4B) w Desc.            & 36.3\% & 69.9\% & 77.8\% & 50.3\% & 7.1\% & 19.2\% & 27.1\% & 13.0\% \\
(4B) w/o Desc.          & 36.7\% & 67.5\% & 75.4\% & 49.7\% & 6.2\% & 17.0\% & 25.2\% & 11.5\% \\
(4B) \dense w Desc.     & 34.2\% & 64.2\% & 74.4\% & 47.3\% & 6.0\% & 15.9\% & 22.5\% & 10.8\% \\
(4B) \add-only          & 30.4\% & 54.2\% & 60.9\% & 41.1\% & 6.0\% & 16.9\% & 24.5\% & 11.4\% \\
(4B) Quality Check off  & 35.0\% & 68.9\% & 78.3\% & 49.2\% & 6.1\% & 18.3\% & 26.8\% & 12.1\% \\
\bottomrule
\end{tabular}
\caption{4B \modelname with default dataset sampling.}
\label{tab:farm-4b-ablations}
\end{table}

\begin{table}[h]
\centering
\renewcommand{\arraystretch}{0.6}
\begin{tabular}{lcccccccc}
\toprule
Method &P@0.1 & P@0.5 & P@1.0 & P@2.0 & R@0.1 & R@0.5 & R@1.0 & R@2.0 \\
\midrule
(4B) w Desc. & 43.2\% & 50.0\% & 55.5\% & 62.1\% & 28.7\% & 66.4\% & 76.6\% & 81.7\% \\
(4B) w/o Desc. & 42.2\% & 49.4\% & 54.9\% & 61.1\% & 27.7\% & 64.7\% & 72.2\% & 78.4\% \\
(4B) \dense w/o Desc. & 40.8\% & 48.2\% & 53.8\% & 60.5\% & 40.0\% & 74.2\% & 82.4\% & 86.4\% \\
(4B) \dense w Desc. & 42.9\% & 49.9\% & 55.5\% & 61.6\% & 35.3\% & 70.4\% & 78.6\% & 82.5\% \\
(4B) \add-only & 38.8\% & 45.6\% & 51.5\% & 58.3\% & 27.9\% & 64.1\% & 72.5\% & 78.8\% \\
(4B) Quality Check off & 38.1\% & 45.3\% & 51.1\% & 57.6\% & 32.5\% & 68.1\% & 77.1\% & 82.1\% \\
\bottomrule
\end{tabular}
\caption{Ablation on MP3D dataset, test18 split, macro-averaged over ObjectNav-21 classes.}
\label{tab:tagmap-4b-ablations}
\end{table}

\begin{table}[h]
\centering
\begin{tabular}{lcccccccc}
\toprule
 & \multicolumn{4}{c}{ScanNet} & \multicolumn{4}{c}{HM3D} \\
\cmidrule(lr){2-5} \cmidrule(lr){6-9}
Method & Acc@1 & R@5 & R@10 & MRR & Acc@1 & R@5 & R@10 & MRR \\
\midrule
(4B) Token Mean & 36.3\% & 69.9\% & 77.8\% & 50.3\% & 7.1\% & 19.2\% & 27.1\% & 13.0\% \\
(4B) Sample Mean & 36.1\% & 67.1\% & 73.0\% & 49.2\% & 6.6\% & 18.4\% & 25.9\% & 12.2\% \\
(4B) Object Mean & 36.6\% & 70.6\% & 78.1\% & 50.8\% & 6.7\% & 19.0\% & 27.2\% & 12.7\% \\
\bottomrule
\end{tabular}
\caption{Scene-graph retrieval summary from both benchmarks, 3D point-in-box (extent-free). Ablating loss reduction techniques. Default dataset weighting is used.}
\label{tab:farm-loss-reduction}
\end{table}

\begin{table}[h]
\centering
\begin{tabular}{lcccccccc}
\toprule
 & \multicolumn{4}{c}{ScanNet} & \multicolumn{4}{c}{HM3D} \\
\cmidrule(lr){2-5} \cmidrule(lr){6-9}
Method & Acc@1 & R@5 & R@10 & MRR & Acc@1 & R@5 & R@10 & MRR \\
\midrule
(4B) Default & 36.3\% & 69.9\% & 77.8\% & 50.3\% & 7.1\% & 19.2\% & 27.1\% & 13.0\% \\
(4B) SQRT.   & 37.3\% & 70.2\% & 78.4\% & 51.0\% & 6.5\% & 18.2\% & 26.6\% & 12.2\% \\
(4B) Uniform & 39.3\% & 71.3\% & 79.2\% & 52.6\% & 6.6\% & 18.8\% & 26.5\% & 12.5\% \\
\bottomrule
\end{tabular}
\caption{Scene-graph retrieval summary from both benchmarks, 3D point-in-box (extent-free). Ablating dataset weighting.}
\label{tab:farm-data-weights}
\end{table}

\begin{table}[h]
\centering
\begin{tabular}{lcccccccc}
\toprule
 & \multicolumn{4}{c}{ScanNet} & \multicolumn{4}{c}{HM3D} \\
\cmidrule(lr){2-5} \cmidrule(lr){6-9}
Method & Acc@1 & R@5 & R@10 & MRR & Acc@1 & R@5 & R@10 & MRR \\
\midrule
(2B) \sparse w Desc.    & 38.3\% & 70.7\% & 77.9\% & 51.8\% & 7.1\% & 19.5\% & 28.1\% & 13.1\% \\
(4B) \sparse w Desc.    & 36.3\% & 69.9\% & 77.8\% & 50.3\% & 7.1\% & 19.2\% & 27.1\% & 13.0\% \\
(8B) \sparse w/o Desc.  & 36.7\% & 67.5\% & 75.4\% & 49.7\% & 5.9\% & 16.5\% & 23.9\% & 11.1\% \\
\bottomrule
\end{tabular}
\caption{Scene-graph retrieval summary from both benchmarks, 3D point-in-box (extent-free). Ablating model size. Default dataset weighting is used. 8B model is trained without descriptions to fit trainings on the hardware we have available.}
\label{tab:farm-model-size}
\end{table}

\begin{table}[h]
\centering
\begin{tabular}{lcccccccc}
\toprule
 & \multicolumn{4}{c}{ScanNet} & \multicolumn{4}{c}{HM3D} \\
\cmidrule(lr){2-5} \cmidrule(lr){6-9}
Img. Res. & Acc@1 & R@5 & R@10 & MRR & Acc@1 & R@5 & R@10 & MRR \\
\midrule
(2B) 112x112 & 31.9\% & 61.5\% & 69.0\% & 44.3\% & 5.0\% & 14.4\% & 20.8\% & 9.5\% \\
(2B) 158x158 & 34.3\% & 66.9\% & 75.0\% & 47.7\% & 5.5\% & 15.6\% & 22.3\% & 10.4\% \\
(2B) 224x224 & 38.3\% & 71.5\% & 79.4\% & 51.9\% & 6.3\% & 18.0\% & 25.9\% & 11.9\% \\
(2B) 316x316 & 38.3\% & 70.7\% & 77.9\% & 51.8\% & 7.1\% & 19.5\% & 28.1\% & 13.1\% \\
(2B) 448x448 & 36.7\% & 70.3\% & 77.1\% & 50.8\% & 7.4\% & 20.1\% & 28.4\% & 13.4\% \\
(2B) 512x512 & 38.0\% & 70.2\% & 76.8\% & 51.3\% & 6.8\% & 19.4\% & 27.8\% & 12.9\% \\ \midrule
(4B) 224x224 & 36.4\% & 70.1\% & 78.3\% & 50.7\% & 6.6\% & 19.0\% & 26.7\% & 12.5\% \\
(4B) 316x316 & 36.3\% & 69.9\% & 77.8\% & 50.3\% & 7.1\% & 19.2\% & 27.1\% & 13.0\% \\
(4B) 448x448 & 37.8\% & 70.5\% & 78.4\% & 52.0\% & 6.7\% & 18.3\% & 26.2\% & 12.4\% \\
(4B) 512x512 & 37.3\% & 69.3\% & 76.4\% & 50.7\% & 6.4\% & 18.4\% & 26.3\% & 12.1\% \\
\bottomrule
\end{tabular}
\caption{Scene-graph retrieval summary from both benchmarks, 3D point-in-box (extent-free). Measuring impact of image max resolution during training and evaluation. Default dataset weighting.}
\label{tab:farm-img-res}
\end{table}

\clearpage
\subsection{Retrieval under the FARM evaluation protocol}
\label{app:sec:farm-more-experiments}

\modelname represents each mapped object by a 3D point and does not directly predict its spatial extent.
In contrast, the original FARM evaluation protocol determines whether a retrieved object is correct using 3D axis-aligned bounding-box intersection over union (AABB IoU), and therefore requires an estimate of object size.
To evaluate \modelname under this protocol, we use Qwen3-VL-4B to estimate the size of the queried object and use this estimate to construct the corresponding 3D bounding box.
The results are reported in \cref{tab:farm-summary-aabb}.

Under the extent-based protocol, \modelname achieves slightly higher Acc@1 than FARM on both ScanNet and HM3D, while FARM performs better on Recall@5, Recall@10, and MRR.
These results indicate that \modelname remains competitive under the original FARM protocol, but also show that extent estimation introduces an additional source of error when the underlying scene representation stores only object positions.
Extending \modelname to predict object extent directly would remove the dependence on an external size estimator and is a promising direction for future work.

\begin{table}[h]
\centering
\begin{tabular}{lcccccccc}
\toprule
 & \multicolumn{4}{c}{ScanNet} & \multicolumn{4}{c}{HM3D} \\
\cmidrule(lr){2-5} \cmidrule(lr){6-9}
Method & Acc@1 & R@5 & R@10 & MRR & Acc@1 & R@5 & R@10 & MRR \\
\midrule
BBQ           & 22.7\% & 35.4\% & 35.9\% & 28.1\% & 3.1\% & 6.5\%  & 7.9\%  & 4.6\% \\
FARM          & 29.4\% & 60.0\% & 70.6\% & 42.8\% & 4.0\% & 11.8\% & 17.9\% & 8.5\% \\
\modelname-4B & 30.0\% & 57.1\% & 64.5\% & 41.4\% & 4.4\% & 11.5\% & 15.7\% & 7.7\% \\
\bottomrule
\end{tabular}
\caption{
Language-grounded retrieval under the original FARM 3D AABB IoU protocol.
Because \modelname predicts object positions without extents, Qwen3-VL-4B is used to estimate object size for bounding-box construction.
}
\label{tab:farm-summary-aabb}
\end{table}

\clearpage
\section{Qualitative Examples}
\label{sec:appendix-qualitative-examples}
We provide additional qualitative examples to better understand the behavior of \modelname during incremental scene mapping. 
In contrast to methods that construct a static representation from a complete observation sequence, \modelname continuously updates a structured object-level map through \add and \edit operations. 
The following examples illustrate both the strengths and limitations of this formulation.

\begin{figure}[h]
    \centering
    \includegraphics[width=\linewidth]{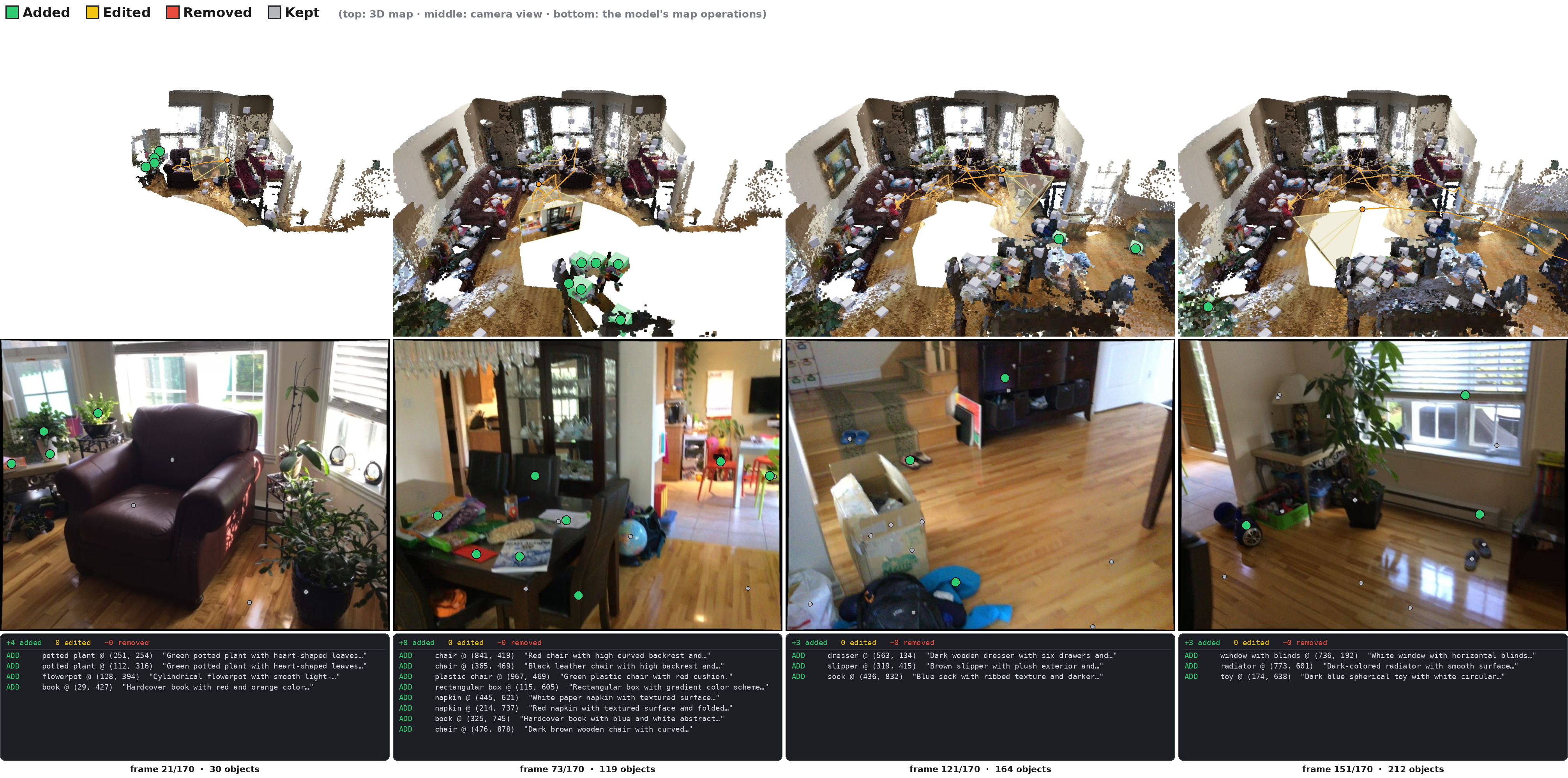}
    \caption{%
    \textbf{Incremental scene expansion.}
    The model successfully adds newly observed objects as they appear in the scene. 
    In frame 21, newly revealed flowers, pots, and books are correctly incorporated into the map. 
    In frame 73, multiple chairs are added with accurate attributes, including color and material, while the paper sheets are incorrectly classified as napkins. 
    In frame 151, the model identifies the hoverboard as a toy but fails to recover the specific category, while correctly adding the window and radiator.
    }
    \label{fig:app:adding}
\end{figure}

\cref{fig:app:adding} shows an example of incremental scene expansion. 
As new regions of the scene become visible, \modelname successfully adds newly observed objects to the map while preserving previously detected entities. 
The model is able to recognize a wide range of object categories and attributes, such as chair colors and materials, demonstrating its open-vocabulary mapping capabilities. 
However, the example also highlights limitations in fine-grained object recognition, where visually similar objects may be assigned an incorrect category.

\begin{figure}[h]
    \centering
    \includegraphics[width=\linewidth]{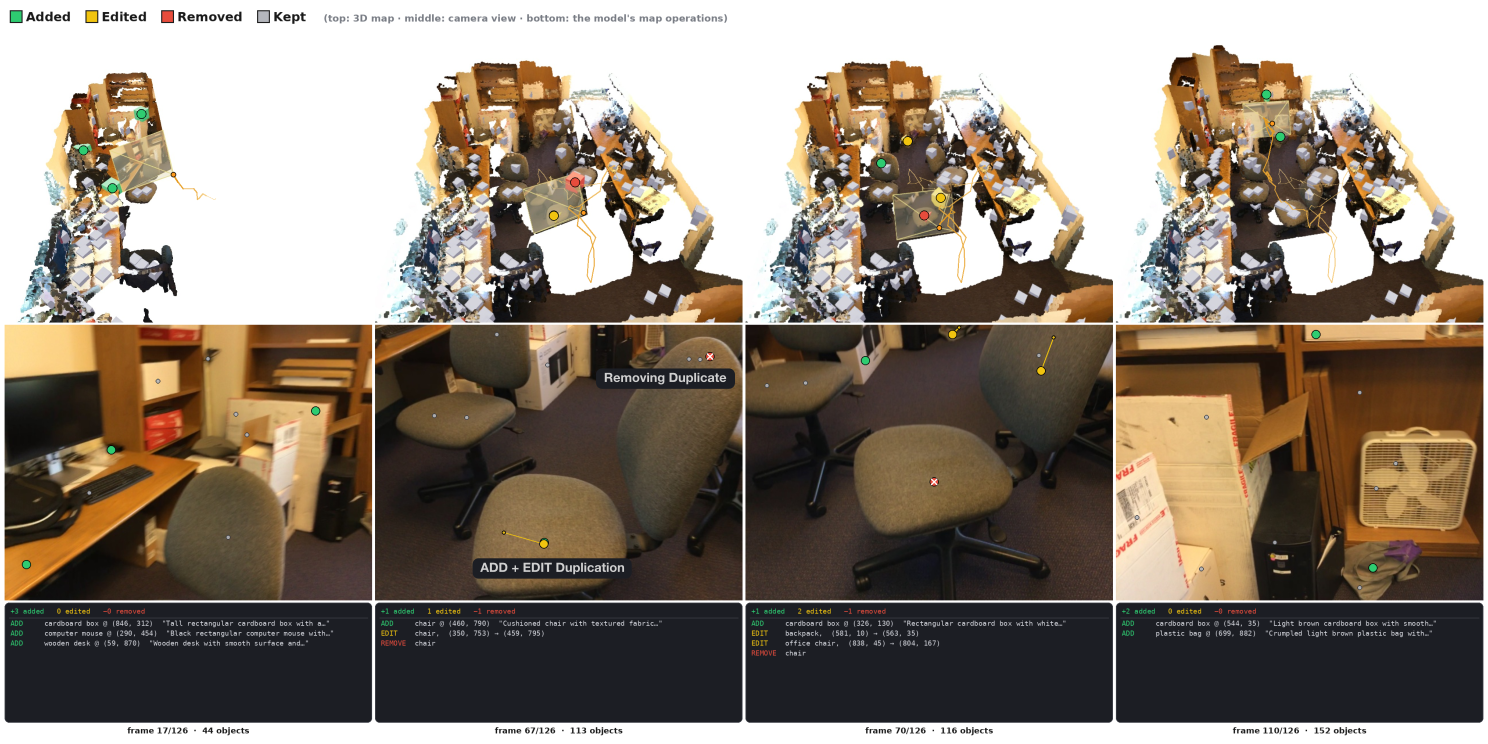}
    \caption{%
    \textbf{Failure cases.}
    The model occasionally produces duplicate objects when an object is both edited and added at the same location. 
    The example also shows successful duplicate removal, where a redundant object introduced in an earlier frame is removed from the map.
    }
    \label{fig:app:failure1}
\end{figure}

\begin{figure}[h]
    \centering
    \includegraphics[width=\linewidth]{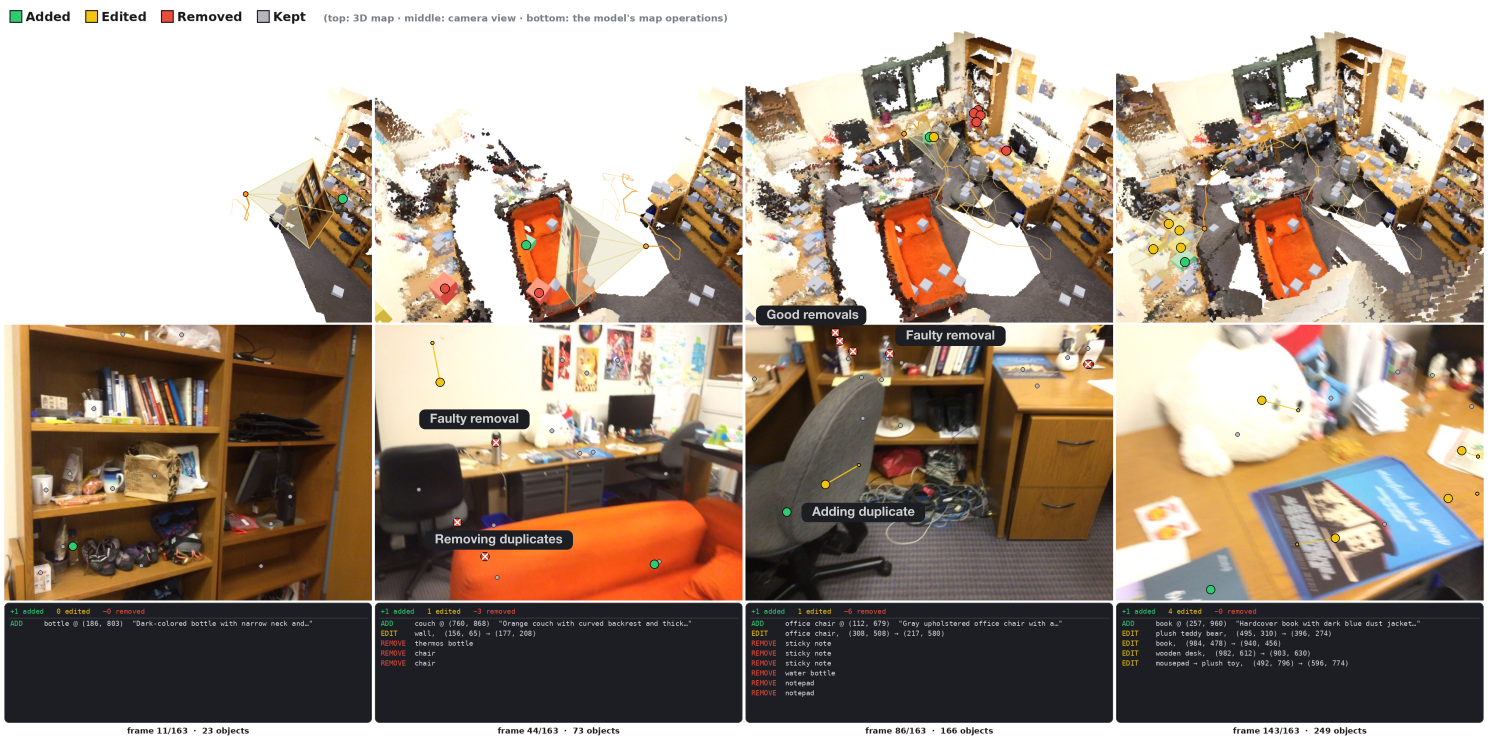}
    \caption{%
    \textbf{Failure cases.}
    The model correctly removes duplicate chairs and incorrectly placed sticky notes, but can also mistakenly remove valid objects, such as the thermos bottle and water bottle.
    }
    \label{fig:app:failure3}
\end{figure}

\begin{figure}[h]
    \centering
    \includegraphics[width=\linewidth]{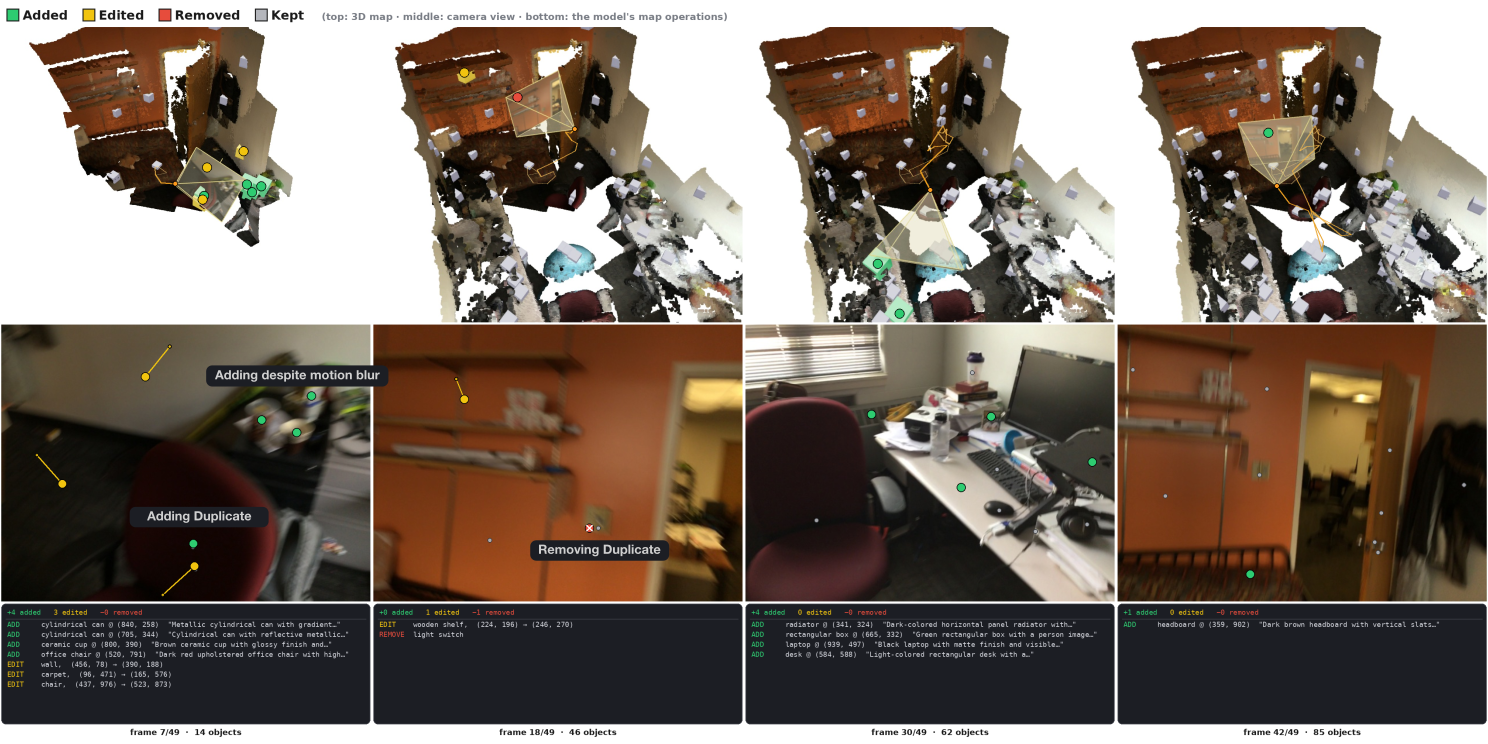}
    \caption{%
    \textbf{Failure cases under challenging observations.}
    Under severe motion blur, the model can still correctly add several objects and remove redundant detections, but may introduce duplicates, such as adding a chair while updating an existing chair position.
    }
    \label{fig:app:failure4}
\end{figure}

Beyond recognition errors, maintaining a consistent object-level map over time introduces additional challenges. 
\cref{fig:app:failure1,fig:app:failure3,fig:app:failure4} show representative failure cases. 
A common failure mode occurs when the model incorrectly associates a new observation with an existing object, resulting in duplicate objects or unnecessary edits. 
Additionally, the model can occasionally apply incorrect removal operations by confusing valid objects with previously introduced errors. 
Nevertheless, these examples also demonstrate successful correction of earlier mistakes, where redundant or incorrectly positioned objects are removed from the map.

\begin{figure}[h]
    \centering
    \includegraphics[width=\linewidth]{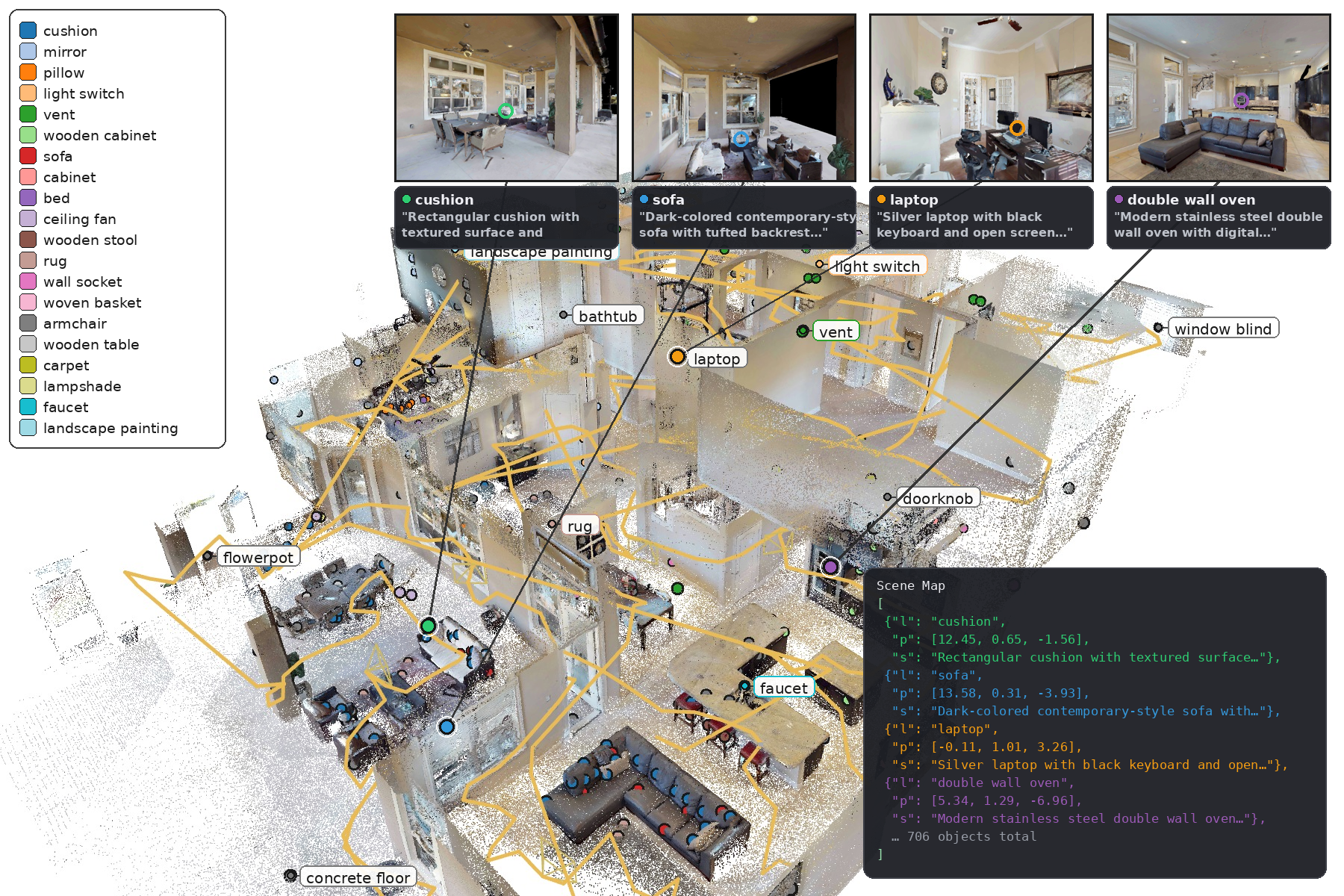}
    \caption{%
    \textbf{Open-vocabulary scene mapping.}
    Visualization of a HM3D scene, demonstrating the ability of the model to represent diverse object categories using open-vocabulary descriptions.
    }
    \label{fig:app:map1}
\end{figure}

\begin{figure}[h]
    \centering
    \includegraphics[width=\linewidth]{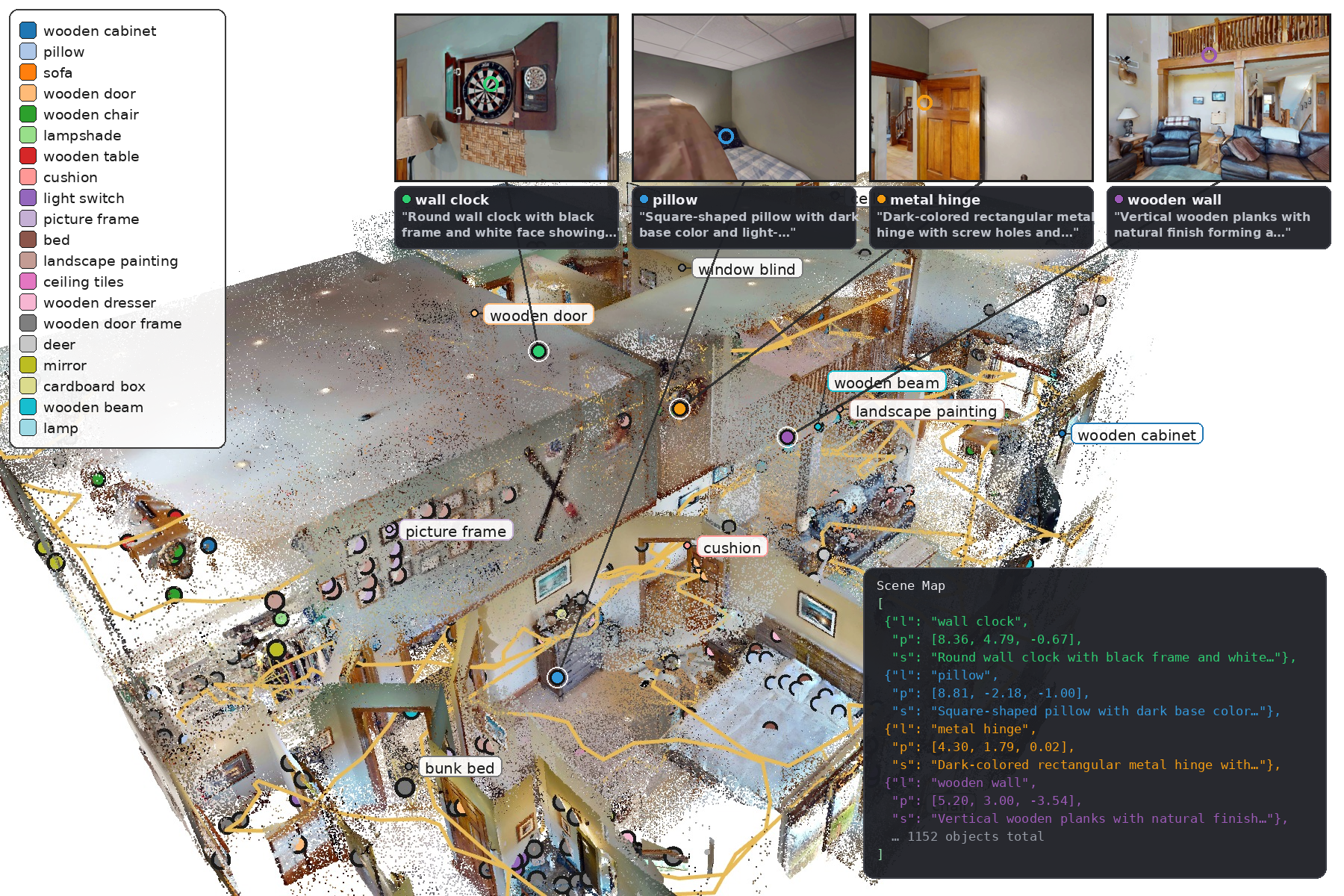}
    \caption{%
    \textbf{Failure case in open-vocabulary recognition.}
    Visualization of a HM3D scene map where the dartboard is incorrectly classified as a wall clock.
    A subset of other objects are highlighted as dots in the map, and a few as insets with a view from the trajectory through the scene.
    }
    \label{fig:app:map2}
\end{figure}

Finally, we visualize complete reconstructed HM3D scene maps in \cref{fig:app:map1,fig:app:map2}. 
These examples demonstrate that \modelname can construct rich open-vocabulary scene representations containing diverse object categories. 
At the same time, they reveal remaining challenges in object-level semantic understanding, where visually similar objects may be confused, such as a dartboard being classified as a wall clock.

\clearpage
\section{Real-world deployment}
\label{app:sec:deployment}

We evaluate whether \modelname can maintain an open-vocabulary scene map online on a mobile quadruped.
We first analyze the deployment trade-offs required for real-time operation on an edge device.
We then evaluate the deployed system in several real-world environments.
Finally, we provide implementation details of the deployed system.

\subsection{Deployment trade-offs}
\label{app:sec:edge}

Deploying \modelname on a mobile robot requires balancing mapping quality against computational throughput.
Unlike the offline experiments in the main paper, an online system must continuously process incoming observations without accumulating an ever-growing queue of images.
We therefore study two aspects of the deployment.
First, we investigate how model quantization affects inference throughput on the robot.
Second, we determine how frequently the scene map needs to be updated as the robot moves.
Together, these experiments define the deployment configuration used for all real-world experiments.

\subsubsection{Model quantization}

We evaluate two model sizes (2B and 4B) using BF16, FP8, and Blackwell-native NVFP4 quantization.
Inference is performed using vLLM on the NVIDIA Jetson AGX Thor mounted on the robot and on an RTX~5090 workstation for reference.
For each configuration, we report both the decoding throughput in generated tokens per second and the sustained scene-map update rate in completed map updates per second.

Table~\ref{tab:edge-tokens} shows that lower-precision inference substantially improves throughput on both platforms.
The gains are particularly pronounced on the Jetson, where inference is increasingly limited by memory bandwidth.
Among all evaluated configurations, NVFP4 consistently achieves the highest sustained update rate while also requiring the smallest memory footprint.
Unless otherwise stated, all subsequent deployment experiments therefore use the 2B model with NVFP4 quantization.

\begin{table}[h]
\centering
\scriptsize
\renewcommand{\arraystretch}{0.9}
\setlength{\tabcolsep}{5pt}
\begin{tabular}{llcccc}
\toprule
 & & \multicolumn{2}{c}{\textbf{Decode speed (tok/s)}} & \multicolumn{2}{c}{\textbf{Update rate (keyframes/s)}} \\
\cmidrule(lr){3-4}
\cmidrule(lr){5-6}
Model & Precision & RTX 5090 & Jetson Thor & RTX 5090 & Jetson Thor \\
\midrule
2B & BF16   & 315 & 50.1 & 2.59 & 0.53 \\
2B & FP8    & 382 ($1.2\times$) & 79.7 ($1.6\times$) & 3.09 & 0.83 \\
2B & NVFP4  & \textbf{460} ($1.5\times$) & \textbf{100.2} ($2.0\times$) & \textbf{3.54} & \textbf{0.86} \\
\midrule
4B & BF16   & 162 & 28.6 & 1.36 & 0.21 \\
4B & FP8    & 232 ($1.4\times$) & 45.1 ($1.6\times$) & 1.91 & 0.33 \\
4B & NVFP4  & \textbf{296} ($1.8\times$) & \textbf{68.0} ($2.3\times$) & \textbf{2.36} & \textbf{0.51} \\
\bottomrule
\end{tabular}
\caption{
Inference throughput for different model sizes and quantization strategies.
Decode speed measures generated tokens per second, while update rate measures completed scene-map updates per second.
}
\label{tab:edge-tokens}
\end{table}

\newpage
\subsubsection{Distance-based scene updates}
\begin{wrapfigure}{r}{0.42\textwidth}
\vspace{-1.0em}
\centering
\includegraphics[width=\linewidth]{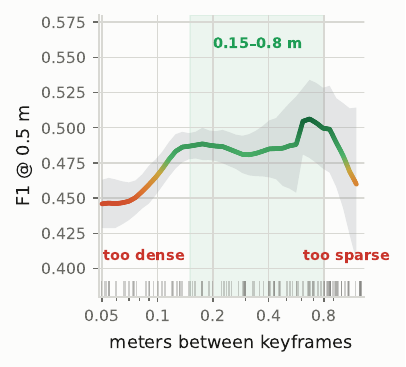}
\caption{
Mapping performance as a function of the travelled distance between processed keyframes.
A spacing of approximately $0.5\,\mathrm{m}$ provides the best balance between redundant observations and scene coverage.
}
\label{fig:spacing}
\vspace{-1.0em}
\end{wrapfigure}
Rather than processing every incoming camera frame, we trigger a scene-map update once the robot has travelled a fixed distance.
This avoids repeatedly processing highly overlapping views while making the update frequency independent of the robot's velocity.
The robot therefore performs scene updates only when sufficient new observations are expected.

To determine a suitable distance threshold, we resample ScanNet trajectories at different spatial intervals and reconstruct the complete scene map from each sampled trajectory.
Figure~\ref{fig:spacing} shows the resulting mapping performance.
Very small spacings introduce highly redundant observations that primarily increase duplicate detections, while large spacings reduce scene coverage and lower recall.
Across the evaluated spacings, approximately $0.5\,\mathrm{m}$ provides the best trade-off between precision and recall.

The update rate of the deployed model and the selected distance threshold together determine the maximum speed at which the robot can maintain online mapping.
If the robot moves faster than the sustained update rate permits, intermediate frames are discarded, increasing the distance between processed keyframes rather than allowing an unbounded processing backlog to accumulate.
Since mapping performance degrades only gradually around the optimum in Figure~\ref{fig:spacing}, occasional frame drops incur only a modest reduction in mapping quality.
Based on these experiments, we deploy the 2B NVFP4 model and trigger scene-map updates every $0.5\,\mathrm{m}$ of robot motion for all real-world experiments.
At a spacing of 0.5 m, the 2B NVFP4 model sustains 0.87 updates/s on the Jetson, corresponding to a nominal operating speed of approximately 0.44 m/s.

\subsection{Real-world experiments}
\label{app:sec:deployment-results}

We evaluate the deployed system in several real-world environments to assess whether \modelname can maintain a useful open-vocabulary scene map under realistic operating conditions.
Unless otherwise stated, all experiments use the deployment configuration established in the previous section, namely the 2B model with NVFP4 quantization and distance-triggered scene updates every $0.5\,\mathrm{m}$.

\subsubsection{Experimental setup}

We evaluate the system in three environments with increasing complexity.
The first is a laboratory environment containing a diverse collection of deliberately arranged objects.
The second is a cluttered office containing many visually similar objects that challenge long-term object association.
The third is an outdoor university campus route spanning several hundred meters and containing buildings, vegetation, street furniture, and construction areas.

Since no ground-truth instance-level 3D annotations are available for these environments, we evaluate the generated maps using a manually constructed object inventory.

\subsubsection{Quantitative evaluation}
The laboratory environment contains many unique objects and achieves the highest recall, with the deployed system recovering 35 of the 44 inventoried objects.
Most missed detections correspond to small or thin objects that are difficult to recognize at the available image resolution.

The office environment is considerably more challenging due to repeated object instances and visually similar furniture.
Here, the primary failure mode is not missing objects but reduced semantic specificity and duplicate objects.
For example, a leather armchair may be described simply as a \emph{dark chair}, while a server rack is represented by several individual computer towers.
Despite these ambiguities, completely hallucinated objects remain rare across all environments.

The outdoor campus demonstrates that \modelname scales beyond individual rooms to larger environments.
Compared to workstation inference, the on-device deployment produces fewer mapped objects due to dropped keyframes when the robot temporarily exceeds the sustained processing rate.
Nevertheless, the resulting scene map remains sufficiently complete for open-vocabulary querying and captures the dominant semantic structure of the environment.

\subsubsection{Qualitative examples}

\begin{figure}[h]
    \centering
    \includegraphics[width=\linewidth]{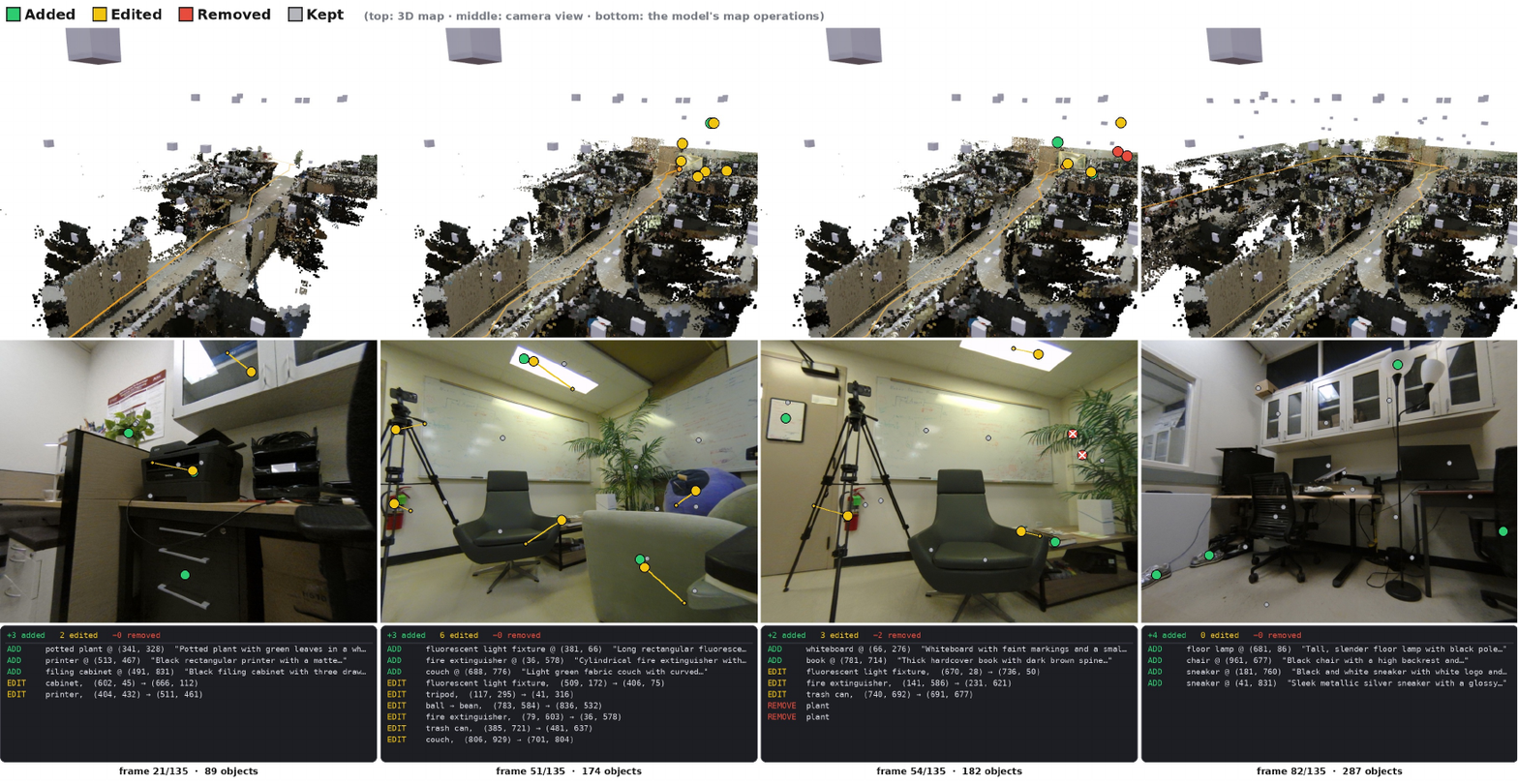}
    \caption{%
    The map generation in the office environment. \modelname runs completely on the quadruped. Noisier pose and depth estimation makes \edit being used more. However, we note more duplicates compared to the ScanNet and HM3D datasets where poses are more accurate.}
    \label{fig:deploy-office}
\end{figure}

\begin{figure}[h]
    \centering
    \includegraphics[width=\linewidth]{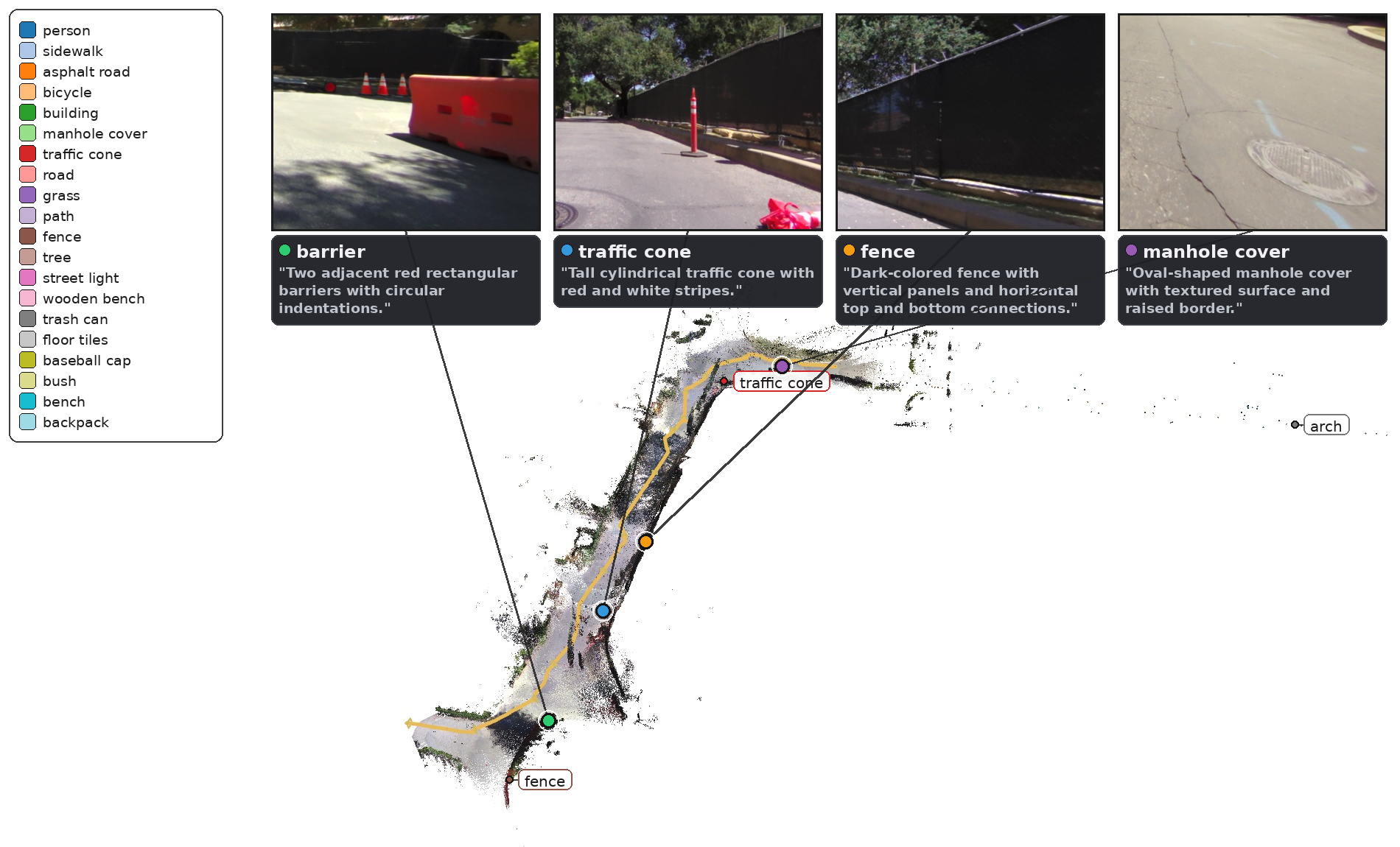}
    \caption{%
    The map generation for a small part of the outdoor campus environment containing a construction zone. \modelname runs completely on the quadruped. }
    \label{fig:deploy-campus-construction}
\end{figure}

\begin{figure}[h]
    \centering
    \includegraphics[width=\linewidth, trim={0 400pt 0 0},clip]{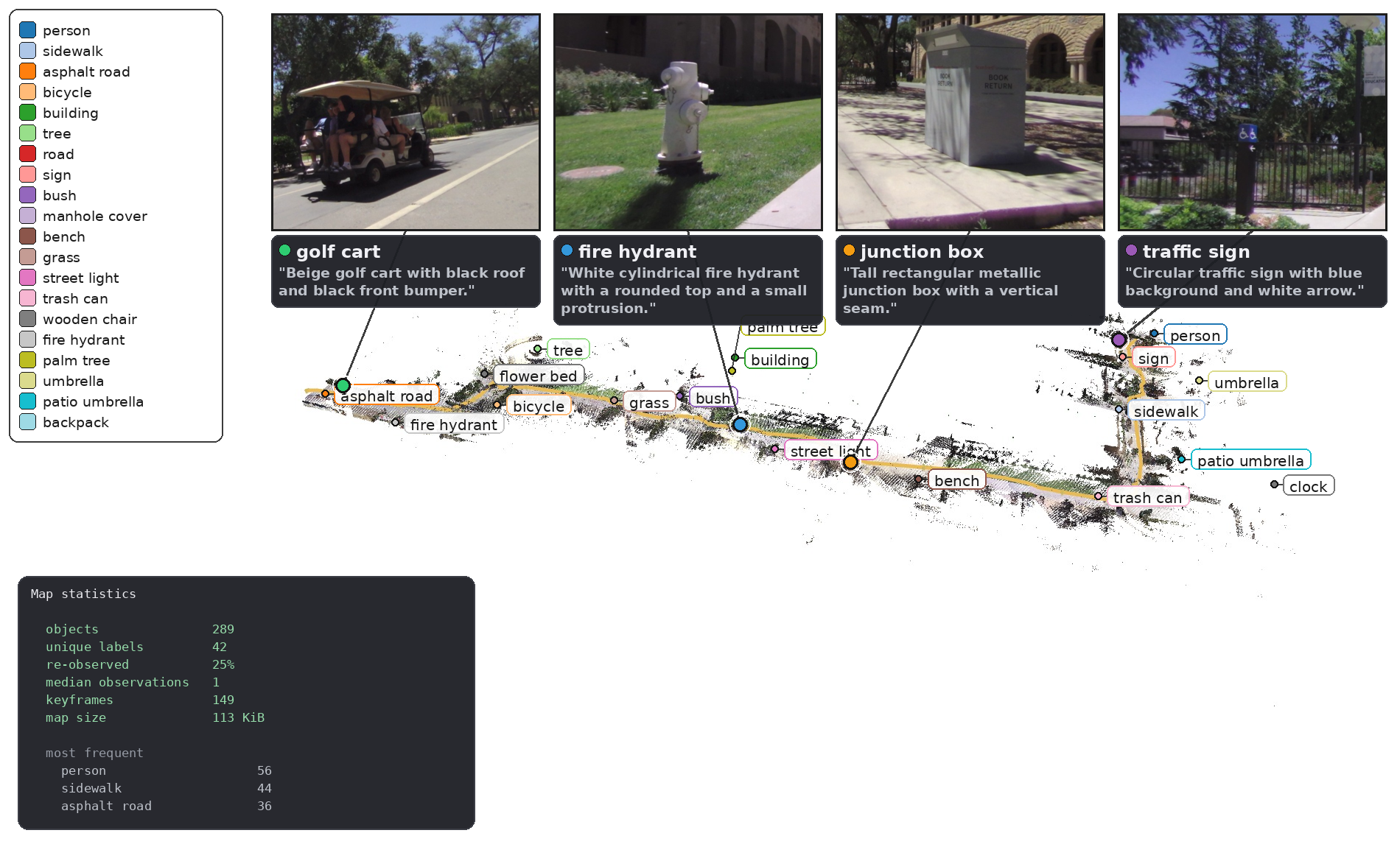}
    \caption{%
    The map generation for a small part of the outdoor campus environment containing a plaza. \modelname runs completely on the quadruped. }
    \label{fig:deploy-campus-plaza}
\end{figure}

\begin{figure}[h]
    \centering
    \includegraphics[width=\linewidth, trim={0 400pt 0 0},clip]{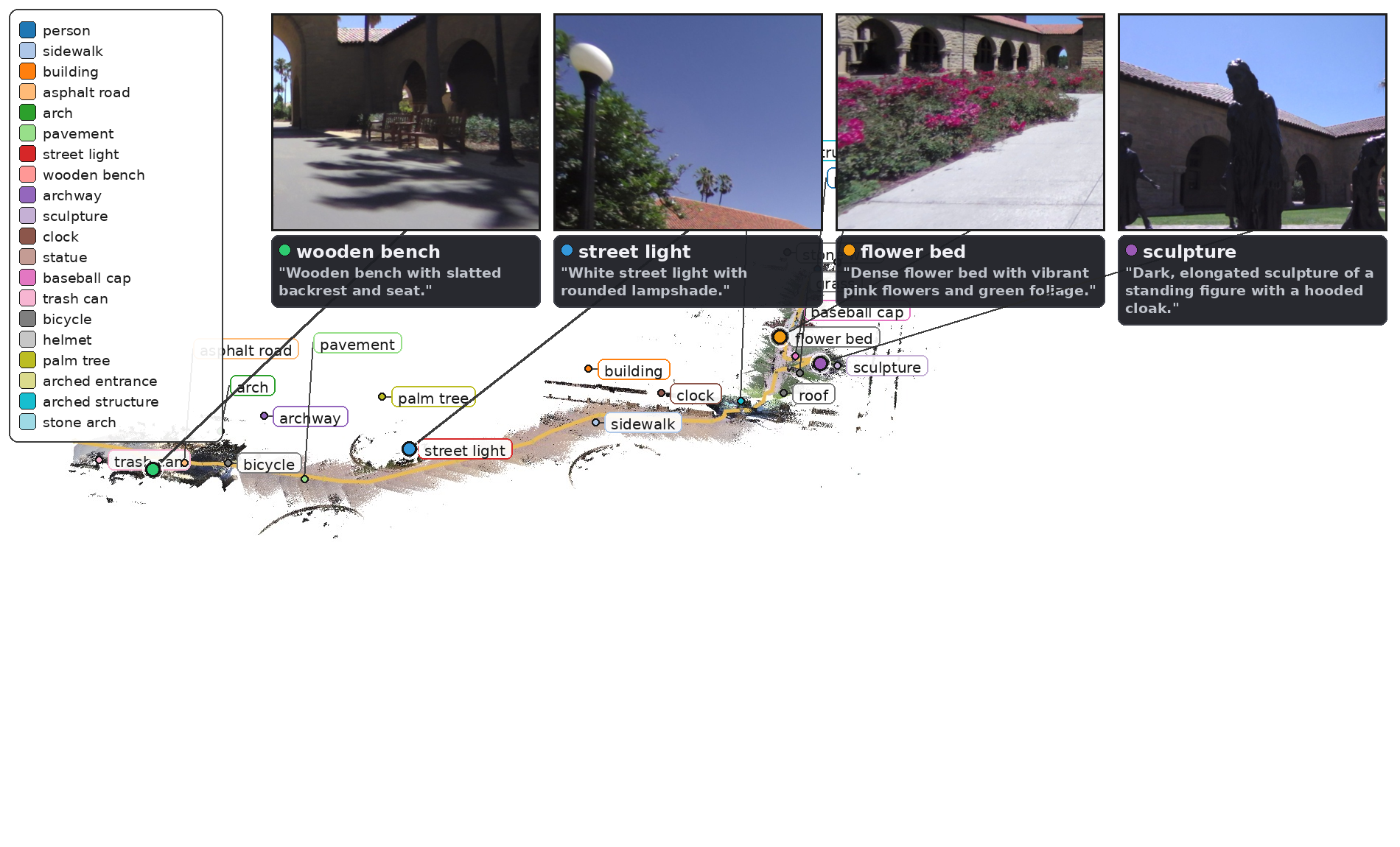}
    \caption{%
    The map generation for a small part of the outdoor campus environment containing a construction zone. 
    \modelname runs completely on the quadruped. }
    \label{fig:deploy-campus-courtyard}
\end{figure}

Figures~\ref{fig:deploy-office}--\ref{fig:deploy-campus-courtyard} present representative scene maps generated during deployment.
The office environment demonstrates that \modelname successfully accumulates observations across multiple viewpoints to construct a coherent semantic map despite noisy odometry and depth estimates.
Compared to the ScanNet and HM3D benchmarks, the real-world deployment exhibits a larger number of duplicate objects, probably due to the increased pose uncertainty of the onboard visual odometry system.

Figures~\ref{fig:deploy-campus-construction}--\ref{fig:deploy-campus-courtyard} show examples from the outdoor campus route.
The generated maps capture diverse semantic elements, including buildings, construction equipment, vegetation, benches, bicycles, and other street furniture, while maintaining a compact textual representation throughout the trajectory.
However, humans moving around tends to add duplicates instead of edited as a new position is detected.
These examples illustrate that the learned mapping policy generalizes beyond indoor environments and can maintain meaningful scene maps over extended real-world trajectories.

\clearpage
\subsection{Implementation details}
\label{app:serving}

The deployment system serves \modelname using vLLM on the NVIDIA Jetson AGX Thor mounted on the robot.
The model runs continuously as part of the robot autonomy stack and receives RGB-D observations together with camera poses estimated by the onboard localization system.
Whenever the robot has travelled $0.5\,\mathrm{m}$ since the previous processed keyframe, the current image and the visible portion of the scene map are forwarded to the model to produce the next scene-map update.

To maximize throughput, we enable vLLM prefix caching since consecutive prompts share a large fraction of their context.
The \add and \edit operations are executed concurrently using a small thread pool, allowing image pre-processing and model execution to overlap.
All deployment experiments use the same inference configuration as the benchmark experiments unless otherwise stated.

The throughput measurements reported in Table~\ref{tab:edge-tokens} are obtained by replaying $36$ ScanNet trajectories through the deployed inference pipeline.
Reported update rates correspond to completed scene-map updates per second, while decode throughput measures generated tokens per second as reported by the vLLM runtime.
Repeating individual trajectories results in only minor variation ($1$--$3\%$), while substantially larger differences are observed between scenes, indicating that throughput is primarily determined by scene complexity rather than runtime variability.

One practical limitation is that no single version of vLLM currently supports all evaluated precisions on the Jetson platform.
BF16 and NVFP4 experiments are therefore performed using vLLM~0.19, while FP8 experiments use vLLM~0.13 due to current kernel compatibility.
Consequently, the reported FP8 throughput slightly understates the performance achievable with newer kernels and should therefore be interpreted as a conservative estimate.

\end{document}